\documentclass[lettersize,journal]{IEEEtran}

\usepackage{microtype}
\usepackage{graphicx}
\usepackage[caption=false,font=normalsize,labelfont=sf,textfont=sf]{subfig}
\usepackage{textcomp}
\usepackage{stfloats}
\usepackage{url}
\usepackage{verbatim}
\usepackage{cite}

\usepackage{hyperref}
\usepackage{tabularray}
\usepackage{threeparttable}
\usepackage{multirow}
\usepackage{adjustbox}
\usepackage{array}
\usepackage{booktabs}
\usepackage{algorithmic}
\usepackage{algorithm}

\usepackage{amsmath}
\usepackage{amssymb}
\usepackage{mathtools}
\usepackage{amsthm}
\usepackage[dvipsnames]{xcolor}

\usepackage[capitalize,noabbrev]{cleveref}

\theoremstyle{plain}

\theoremstyle{definition}

\theoremstyle{remark}

\begin{document}

\title{UniGIO: Unified Generative Global In-situ Weather Modeling from Spatiotemporal Incomplete Observations}

\author{Songru~Yang,
  Zili~Liu$^\star$,
  Tao~Han,
  Ben~Fei,
  Lei~Bai,
  Chang~Liu,
  Zhengxia~Zou$^\star$ ~\IEEEmembership{Senior Member,~IEEE},
  Xiangyang~Ji ~\IEEEmembership{Member,~IEEE},
  Wanli~Ouyang,
  Zhenwei~Shi ~\IEEEmembership{Senior Member,~IEEE},
  \thanks{
  This work was supported in part by the National Natural Science Foundation of China under Grants U24B20177, 62125102, U25A20401, and 62471014.}
  \thanks{
    Songru Yang, Zhenwei Shi and Zhengxia Zou are with the Department of Aerospace Intelligent Science and Technology, School of Astronautics, Beihang University, and with the Key Laboratory of Spacecraft Design Optimization and Dynamic Simulation Technologies, Ministry of Education, Beihang University, Beijing 100191, China.

    Zili Liu, Tao Han, Lei Bai, Wanli Ouyang are with Shanghai Artificial Intelligence Laboratory, Shanghai 200232, China.

    Ben Fei is with Department of Information Engineering, The Chinese University of Hong Kong, Hong Kong 999077, China

    Chang Liu, Xiangyang Ji are with Department of Automation, Tsinghua University, Beijing 100084, China
  }
\thanks{Corresponding authors: Zili Liu (liuzili@pjlab.org.cn), Zhengxia Zou (zhengxiazou@buaa.edu.cn).}}

\markboth{IEEE Transactions (Draft)}{Yang \MakeLowercase{\textit{et al.}}: UniGIO}

\maketitle

\begin{abstract}
  Global In-situ Observation (GIO) provides fine-scale, direct records of the global weather system from sparse point stations, making it an indispensable source for capturing localized and transient dynamics beyond the reach of satellite gridded data, and playing a critical role in key fields such as numerical weather prediction, disaster prevention, and agriculture. However, GIO exhibits strong spatiotemporal incompleteness, severely impairing accurate and real-time in-situ weather modeling. Unlike existing methods waiting for completed AI-ready data with extra introduced errors, in this work, we explore UniGIO, a novel generative framework for directly modeling global in-situ weather dynamics from native incomplete GIO. By generating missing data from observed ones annotated by masks, it unifies the coexisting forecasting, imputation, and generation under arbitrary missing ratios. Between the missing and observed, UniGIO captures station and region level complementarity through the Observation Mixer and Event Aligner, which diffuse discrete observations into continuous spaces where weather processes naturally span multiple stations. We further establish temporal dependencies with pattern shifts using the Adaptive Temporal Mixer, and track extreme events in chaotic local weather systems through a Mixture-of-Experts structure. Steady and extreme events are adapted in decoder by a Local Refiner. Extensive experiments on the up-to-date largest global station weather dataset Weather-5K validate its SOTA performance with 11\%, 12\%, and 5\% advantages on accuracy, fidelity, and extreme event capture, delivering a novel holistic solution for weather modeling in GIO networks.
\end{abstract}


\section{Introduction}
Global In-situ Observation (GIO) provides timely and accurate direct observations of the meteorological variables at the finest scale. \cite{WMO2018GOS}. Depicting local weather systems' evolution, volatility, and extreme events, in-situ weather modeling underpins the assimilation and validation in numerical weather prediction, and serves as a critical tool for various social sectors requiring efficient regional weather services, such as aviation, agriculture, and disaster monitoring \cite{zhu2026foundations}. Existing methods have already supported weather modeling for key scenarios such as the Winter Olympics \cite{wu2023interpretable}.

\begin{figure}[!t]
  \centering
  \includegraphics[width=\linewidth,height=0.6\linewidth]{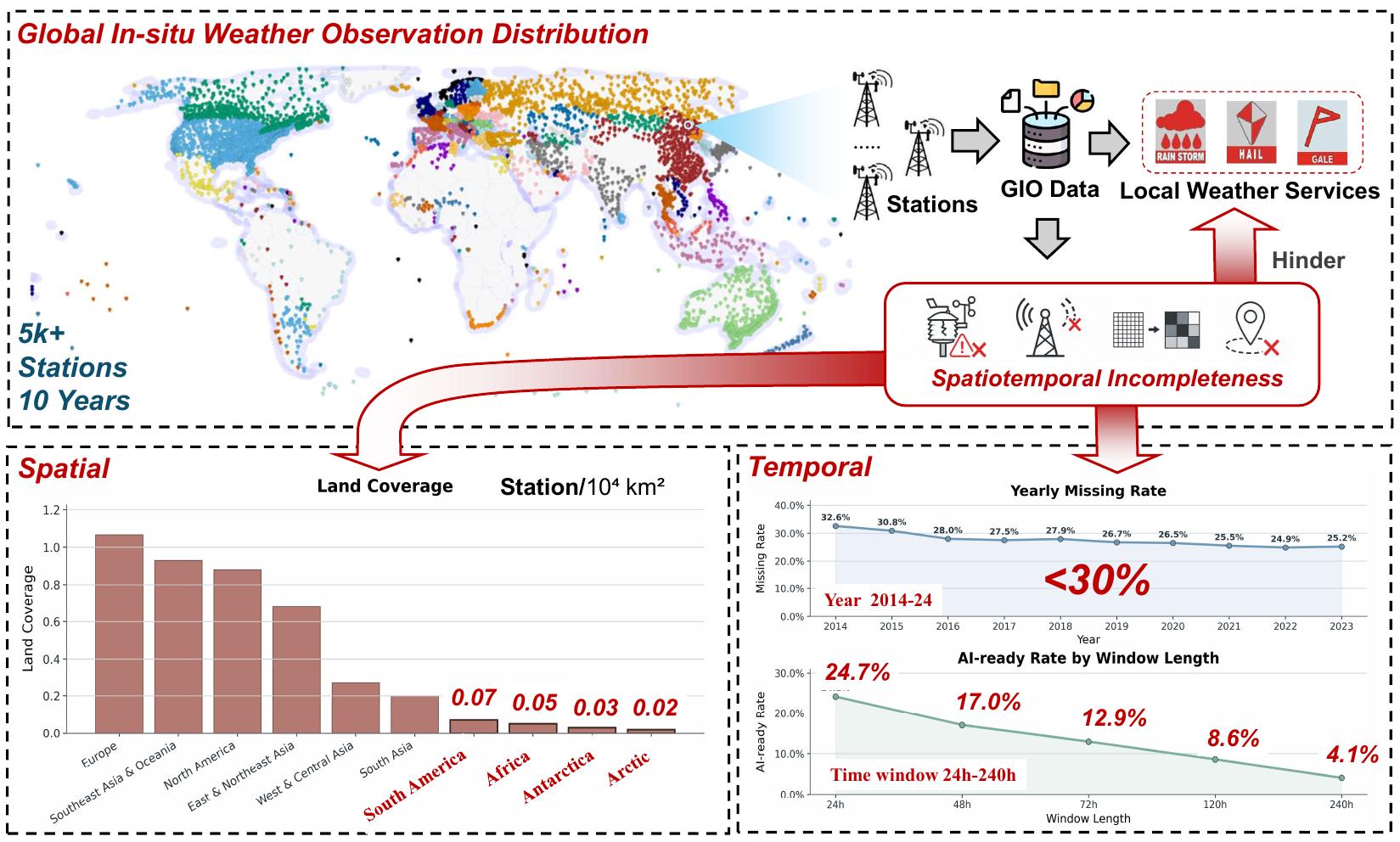}
  \vspace{-10pt}
  \caption{Spatiotemporal incompleteness in GIO makes it difficult to model local weather systems and detect extreme events at the finest scale. We analyzed the station density, annual missing rate, and the AI-ready rate to demonstrate its severity. AI-ready rate: The ratio of non-missing to all samples when station data is spliced in windows.}
  \vspace{-7pt}
  \label{intro_fig1}
\end{figure}

\begin{figure*}[!t]
  \centering
  \includegraphics[width=\textwidth,height=0.45\textwidth]{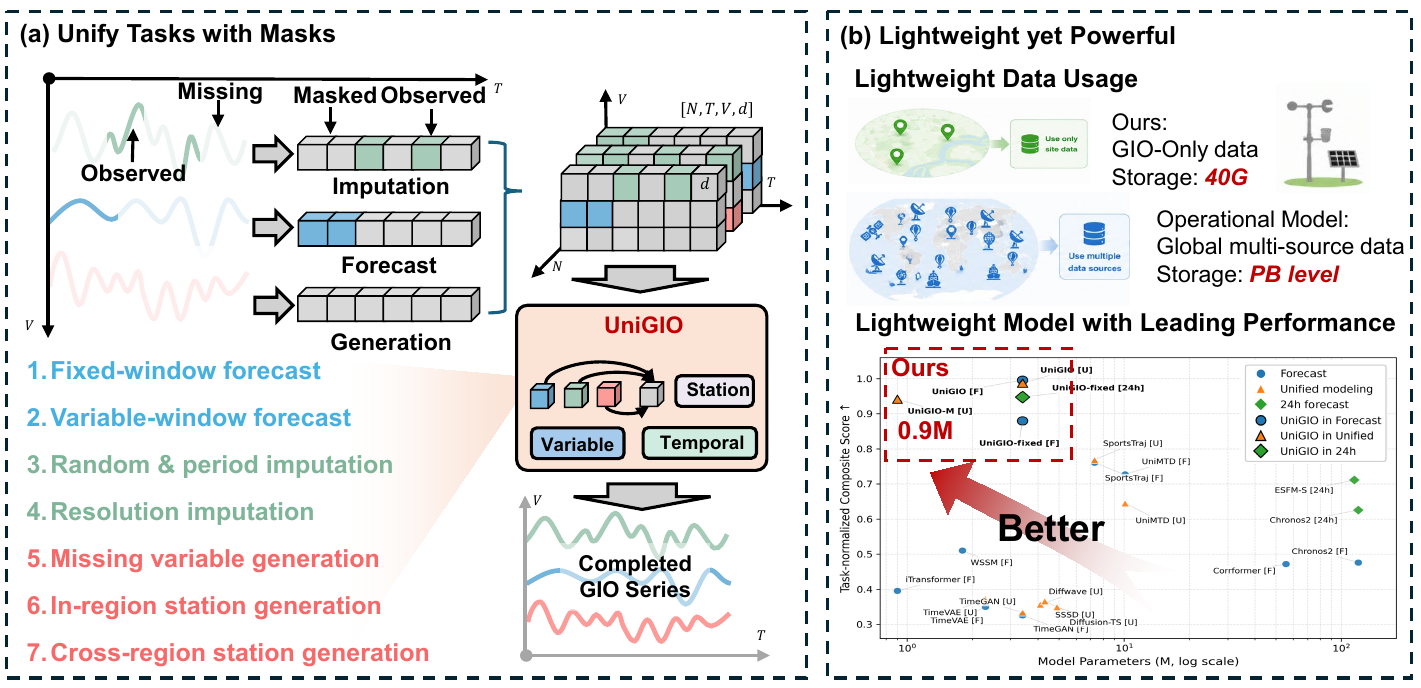}
  \vspace{-10pt}
  \caption{(a) Mask-driven observation-to-missing generation, (b) UniGIO achieves leading performance while maintaining lightweight in data and model scale}
  \vspace{-7pt}
  \label{intro_fig}
\end{figure*}

GIO naturally provides lightweight sub-grid, point-level observations of instantaneous atmospheric dynamics. Although mainstream heavy grid-based reanalysis and satellite data \cite{bi2022pangu, lam2023learning, price2025probabilistic, vaughan2024aardvark,ozdemir2026earth} offer global dense coverage, they rely on interpolation to sub-grid scales, which leads to extreme event and transient signal loss with data inefficiency. GIO serves as a primary source for local weather services and plays a critical role in bridging global forecasts toward local dynamics, providing a foundation for fine-scale weather modeling \cite{kalnay2003atmospheric, bauer2015quiet}.

Nevertheless, GIO exhibits strong spatiotemporal incompleteness, a common and challenging issue in real-world scenarios \cite{cao2018brits,tashiro2021csdi}. In Fig. \ref{intro_fig1}, for Weather-5K dataset \cite{han2024weather} across 5k+ global in-situ stations in 10 years, the data missing rate is 27.5\%, yet AI-ready samples with complete 24h-120h window are less than 25\%-10\%, manifested in three patterns: (1) data missing due to sensor failures or transmission loss; (2) uneven spatial data distribution due to geographical constraints; (3) Temporal resolution differences among various stations \cite{WMO2018GOS,WMOGBON}. This inherent, inevitable data gap leaves the GIO system ill-suited to cutting-edge AI frameworks.

Existing methods in GIO systems remain limited. Forecasting methods \cite{wu2023interpretable,yang2025wssm} often bypass observation gaps by relying on non-learnable pre-processing or cascaded models to construct AI-ready complete series, which introduces data distortion, time delays, cascading errors, and sensitivity to task variations such as different prediction windows. time series generation offers a possible way to handle different missing patterns \cite{yuandiffusion,yang2025unified}, but often yields over-smoothed or noisy results. Recent observation foundation models \cite{gong2026earth,vaughan2024aardvark,ozdemir2026earth} theoretically support arbitrary-location modeling, yet they operate at the global scale and still treat GIO modeling as latent interpolation, leading to averaged results, redundant multi-source inputs, and unaffordable computation for in-situ weather services.

Beyond accuracy degradation, incomplete GIO can obscure local extreme weather events, which account for up to 90\% of global extremes and have caused 70\% of 2 million deaths and 3.64 trillion in losses over the past 50 years \cite{wmo2021atlas,undrr2022gar}. The risk is amplified in Africa, where weather-related mortality is the highest, station density is only $7e^{-6}/km^2$ ($\sim1/84$ of Germany), and only 22\% of stations meet the GBON standard \cite{wmo2019strategic,wmoSOFF,rogers2019weathering,WMOGBON}. As stations are more likely to fail under extremes, it is urgent to capture abrupt local events and hidden risks from incomplete observations \cite{wmo2024guide,wmo2021atlas}.

These unresolved challenges motivate us to propose a novel unified GIO-centric modeling framework named UniGIO, operating directly on interpolation-free distributed weather stations, emphasizing 3 insights: (1) GIO complementarity; (2) Extreme patterns; (3) Lightweight. UniGIO regards observation gaps as coexisting forecasting, imputation, and generation tasks, and unifies them within a mask-driven generation from observation to missing parts as Fig. \ref{intro_fig} (a). Specifically, for complementarity, an Observation Mixer is designed to complement missingness in series observations and continuous regions where weather events evolve; an Event Aligner is added to alleviate the delays of the same event reaching different stations. To handle extreme mutations, an Adaptive Temporal Mixer is designed, leveraging an uncertainty-weighted state space model to establish robust temporal dependencies, and a Pattern Decoupled MoE to group global affinity and separate local heterogeneity. A Local Refiner is added to maintain local continuity. For lightweight design, UniGIO is efficient in both data and model scale. It is trained directly on 40G sparse GIO data instead of PB-level dense global observations, and can be compressed to fewer than 1M parameters while still achieving leading performance, as Fig. \ref{intro_fig} (b). All tasks are unified within a conditional VAE (CVAE) \cite{sohn2015cvae}, which quantifies extreme event risks through variance, supporting its assessment and controllable generation.

We conducted quantitative and qualitative experiments under both native and curated spatiotemporal incompleteness to benchmark the in-situ weather modeling task on the up-to-date largest GIO dataset Weather-5K \cite{han2024weather}, recording hourly temperature, dew point, wind direction, wind rate, and sea-level pressure from 5k+ surface weather stations worldwide, covering 6 continents and 10 years. UniGIO uniformly handles 7 tasks, including fix/variable-window forecasting, random\&period/resolution imputation, variable/in-region/cross-region station generation, and outperforms baselines by an average 9\% across 7 metrics, respectively 11\%, 12\%, and 5\% improvement on accuracy, fidelity, and extreme event capture. Our contributions are as follows:

\begin{itemize}
  \item We propose UniGIO, a novel unified framework for versatile and lightweight global in-situ weather modeling from incomplete GIO through an observation-to-missing generation to advance all-time in-situ service, even when real-world stations are absent or inoperable.

  \item We design Observation Mixer, Event Aligner to capture station and region complement; Adaptive Temporal Mixer, Local Refiner are designed for extreme pattern adaptation; CVAE is specially adopted for observation constraints and evaluate extreme risks.

  \item We benchmark across 5k+ surface weather stations worldwide, covering a 10-year period to establish strong baselines for in-situ weather modeling. Extensive experiments validate our consistent and outstanding performance across tasks, regions, and mask formats.
\end{itemize}

\section{Related Work}
\subsection{Data-driven Weather Prediction}
Since 2022, the AI and atmospheric science communities have seen a rapidly growing interest in data-driven Numerical Weather Prediction (NWP) models like Pangu-Weather \cite{bi2022pangu}, GraphCast \cite{lam2023learning}, GenCast \cite{price2025probabilistic}, and FuXi \cite{chen2023fuxi} et al running on gridded reanalysis data (e,g., $0.25^\circ$ and $0.09^\circ$ resolution). In the case that these large-scale, region-averaged methods may not match the station level weather systems and can ignore highly destructive local extreme weather, some initial attempts, such as Corrformer \cite{wu2023interpretable} and WSSM \cite{yang2025wssm}, have treated station weather forecasting as an independent task and achieved promising results. However, these methods operate on spatiotemporal complete data, remaining a gap to the incomplete GIO, which greatly limits their real-world applications.

Recently, spatiotemporal weather foundation models \cite{gao2022earthformer,nguyen2023climax} and observation foundation models \cite{gong2026earth,vaughan2024aardvark,ozdemir2026earth} have shown the potential to unify station and grid weather modeling at arbitrary locations. Nevertheless, this paradigm remains poorly suited to GIO. It relies on massive multi-source data and intensive computation for global-scale autoregressive forecasting, which is incompatible with in-situ weather applications. Also, although they implicitly incorporate data assimilation into grid-like structured latent spaces to support observation inputs, their GIO forecasts are still derived from interpolation over smoothed latent fields. As a result, they remain fundamentally close to grid-based forecasting and inherit its key limitations. These challenges indicate that a unified weather modeling framework tailored to GIO is not only indispensable for practical in-situ services, but also essential for bridging the gap in the grid-station and global-local weather modeling.

\subsection{Multivariate Time Series Forecasting and Imputation}

Forecasting and imputation are two fundamental tasks in multivariate time series modeling. Forecasting aims to infer future dynamics from historical observations, while imputation focuses on recovering missing values from partially observed sequences. In early studies, CNN-based methods \cite{lim2021time} were adopted to extract local temporal patterns, and RNN-based methods \cite{hewamalage2021recurrent} were widely used to model sequential dependencies. Recently, Transformer-based architectures \cite{wang2025lightgts} have enabled more flexible long-range correlation modeling; this trend is consistent with non-local attention \cite{wang2018nonlocal}, hierarchical vision transformers \cite{liu2021swin,han2023vitsurvey}, and deformable attention \cite{zhu2021deformabledetr}. State space models \cite{ma2025timepro} further introduce temporal inductive biases and recurrent-like computation to construct efficient and robust long-term dependencies. Imputation is often regarded as a preprocessing step for slightly incomplete data. Classical non-learnable methods, such as interpolation \cite{noor2015comparison}, regression \cite{burgette2010multiple}, and EM algorithms \cite{rahman2016missing}, are simple and efficient, but they usually rely on smoothness or distributional assumptions and struggle to recover complex non-stationary sequences. To improve imputation quality, recent learning-based frameworks \cite{liu2019naomi, xu2023uncovering} introduce neural structures such as recurrent networks, attention mechanisms, and graph neural networks to exploit temporal continuity and cross-variable dependencies.

More recent efforts attempt to perform forecasting under missing observations \cite{yu2025merlin, huang2025std}, indicating the growing need for models that are robust to incomplete inputs. However, most existing methods still treat forecasting and imputation as separate tasks, where imputation is either performed as an independent preprocessing stage or forecasting is designed for a specific missing pattern. This separation limits their applicability to Global In-situ Observation (GIO), where missingness is not slight or fixed but arbitrarily coexists across variables, stations, regions, and forecast horizons. From a broader learning perspective, current data synthesis and reconstruction methods have shown that imperfect observations and distributions can substantially impair generalization and reliability \cite{pan2010transfer,nikolenko2021synthetic,wang2021superresolution}. To solve these problems, our work formulates these incomplete patterns with unified observation mask and models forecasting, imputation, and generation within a single framework.

\subsection{Generative Time Series Modeling}

Generative time series modeling aims to synthesize realistic time series by learning the underlying data distribution, often generating sequences from random noise. Early studies mainly explored GAN-based \cite{yoon2019time} and VAE-based \cite{desai2021timevae} frameworks, which respectively model temporal distributions through adversarial learning and latent-variable reconstruction. More recently, diffusion models \cite{yang2023diffusionsurvey,croitoru2023diffusionvision,cao2024diffusionsurvey} have become a dominant paradigm due to their stable training and strong distribution modeling capability in related applications. These generative paradigms \cite{pan2021gansurvey,gui2023ganreview,bondtaylor2022deepgen} focusing on data distribution learning, laid the foundation for task unification. By controlling generation conditions, generative methods can naturally support forecasting, imputation, and generation simultaneously under arbitrary masks, providing a flexible formulation for incomplete time series modeling.

Despite this natural unification capability, most existing generative time series methods are still primarily designed for unconditional or weakly conditional generation. They focus on fitting the global data manifold, which becomes problematic for Global In-situ Observation (GIO), where the global weather manifold is often dominated by smooth evolution, whereas local dynamics are frequently affected by strong extreme noise. Simply landing on this manifold may lead to over-smoothed predictions that miss local extremes; worse, under sparse and noisy conditions, models often fit spurious noise instead of real dynamics, resulting in uncontrolled fluctuations.

We empirically apply advanced generative time series methods, such as diffusion models, to GIO and find that their conditional information is often weakly exploited or even largely discarded, resulting in degraded performance. This indicates that available observations serve as critical anchors for preserving realistic local states and constraining physically plausible generation. Motivated by this, UniGIO adopts a CVAE that tightly anchors valid observations to the latent prior mean and explicitly leverages the variance to model the uncertainty of extreme events, thereby balancing smooth temporal evolution and localized abrupt mutations in a decoupled manner.

\section{Method}

\begin{figure*}[!t]
  \centering
  \includegraphics[width=\textwidth,height=0.5\textwidth]{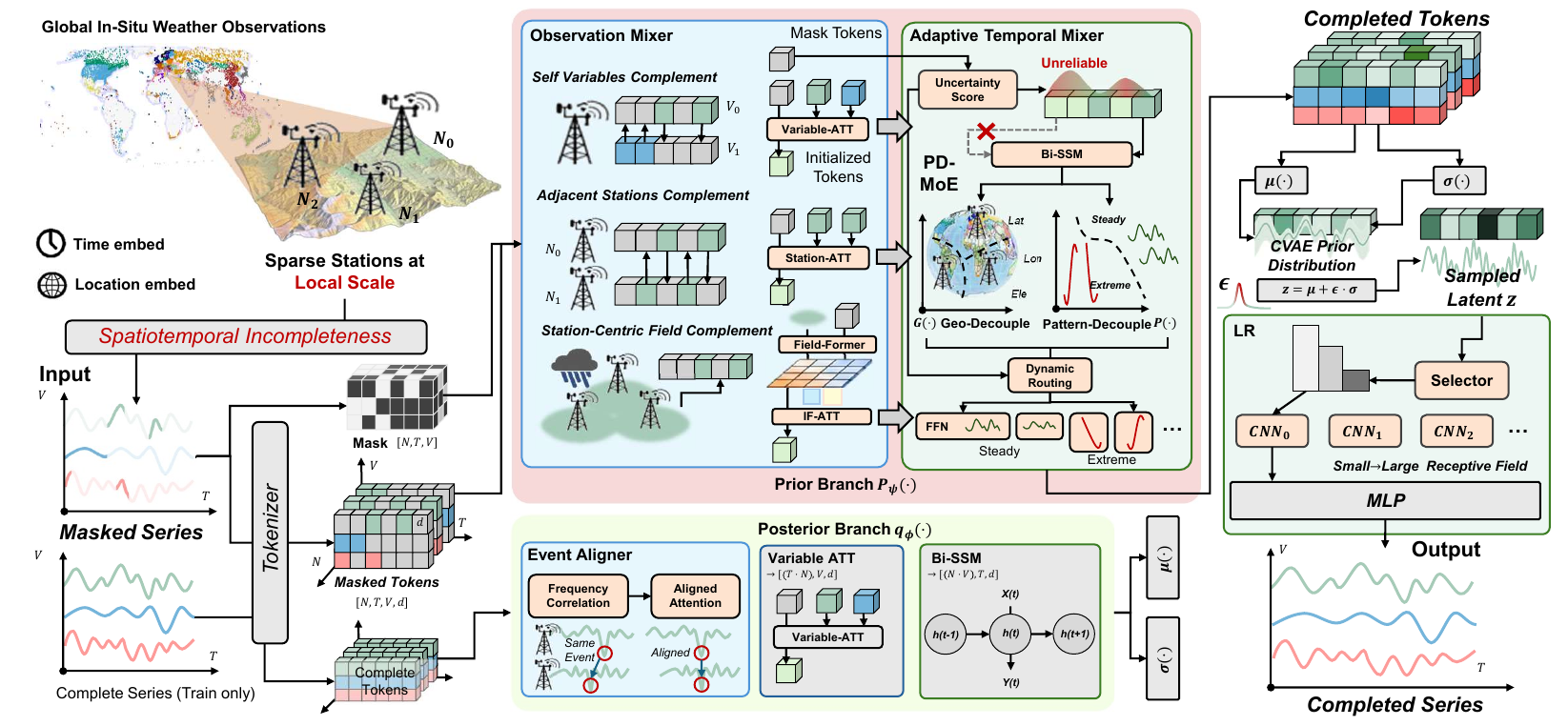}
  \vspace{-17pt}
  \caption{Overall pipeline of UniGIO. The framework is organized as a lightweight CVAE, where key designs for GIO complementation are marked in \textcolor{blue}{blue} and key designs for extreme pattern capture are marked in \textcolor{green}{green}.}
  \label{method_fig}
\end{figure*}

\subsection{Task Formulation}
\label{mask form}
To unify multiple in-situ weather modeling tasks within a single framework, we use an observation mask to represent different incomplete patterns. Within the time window $T$, weather series with $N$ adjacent stations and $C$ variables can be denoted as $\mathbf{X}\in\mathbb{R}^{N\times T\times C}$. The observation mask $\mathbf{M}_n\in\mathbb{R}^{T\times C}$ for $n$-th station can be obtained as:

\begin{equation}
  \begin{aligned}
    \mathbf{M}_{n,t,c} \in \{0, 1\},\ & \text{s.t.}\ \mathbf{M}_{n,t,c} = \mathbf{1}(X_{n,t,c} \text{ is observed}), \\
    &\forall t \in \{1, \dots, T\}, c \in \{1, \dots, C\}
  \end{aligned}
\end{equation}

Given weather series $\mathbf{X}$, observation mask $\mathbf{M}$, and metadata $\mathbf{I}\in\mathbb{R}^{T\times d_m}$, our task is to train a model $f_\theta(\cdot)$ to generate the missing data from the incomplete input $\mathbf{X}_M$, which can be formulated as a self-supervised learning task:

\begin{equation}
  \min_\theta\mathcal{L}(\mathbf{X},f_\theta(\mathbf{X}_M,\mathbf{M},\mathbf{I}))
\end{equation}

Through the various designs of observation masks, we can cover a series of tasks from broad to strict. Taking the forecast mask with look-back window size $i$ as an example:

\begin{equation}
  \mathbf{M}_{n,t,c} = 1(t \le i), \forall t \in \{1, \dots, T\}, c \in \{1, \dots, C\}
\end{equation}

If $i$ is randomly sampled, variable-windows can be achieved; fixed-windows are implemented by fixed $i$.

In this work, we have identified 7 coexisting forecasting, imputation, and generation tasks: 1) fixed-window forecasting, 2) variable-window forecasting, 3) random\&period imputation, 4) resolution imputation, 5) variable generation, 6) in-region station generation, and 7) cross-region station generation. Mask formulations are as follows:

\textbf{Fixed-window/Variable-window Forecasting}:
\begin{equation}
\mathbf{M}_{n,t,c} = 1(t \le i)
\end{equation}
If $i$ is randomly sampled, variable-window forecasting can be achieved; fixed-window forecasting is implemented by fixed $i$. Notably, the missing future in the forecast task should not exhibit complementarity. To achieve this, $N$ stations are aligned to Station 0 to determine the minimum mask length.

\textbf{Random\&period Imputation}:
\begin{equation}
\mathbf{M}_{n,t,c} =
\begin{cases}
0, & (t,c) \in \mathcal{T}_{\text{point}} \ (\mathbb{P}=p_{\text{point}}) \\
&\lor t \in \bigcup_{l=1}^L [a_l, b_l] \\
1, & \text{otherwise}
\end{cases}
\end{equation}
where $\mathcal{T}_{\text{point}} \subseteq \{(1,1),\dots,(T,C)\}$ is the set of isolated points randomly selected from time window $T$ with $C$ variables, each point is selected with probability $p_{\text{point}}$ for random imputation. $\bigcup_{l=1}^L [a_l, b_l]$ is the union of randomly sampled $L$ segments, and the length of each segment $[a_l, b_l]$ is also randomly sampled for period imputation. Notably, considering the independence between variable-specific sensors, we randomly trigger variable unmasking to remove mask for randomly select variables within random sampled data blocks.

This masking strategy accounts for scenarios where stations or sensors malfunction over a period of time or data is randomly unavailable.

\textbf{Resolution Imputation}:
\begin{equation}
\mathbf{M}_{n,t} =
\begin{cases}
0, & t \not\equiv 0 \pmod{S} \\
1, & t \equiv 0 \pmod{S}
\end{cases}
\end{equation}
where $S$ is randomly sampled resolution step size, time steps $t$ not an integer multiple of $S$ are masked to simulate variations in the temporal resolution.

\textbf{Variable Generation}
\begin{equation}
\mathbf{M}_{n,c} =
\begin{cases}
0, & c = c^* \\
1, & c \neq c^*
\end{cases}
\end{equation}
where $c^*$ is a single variable to mask randomly chosen from $C$. This reflects the variation in sensor types for different variables among stations.

\textbf{In-region/Cross-region station generation}
\begin{equation}
\mathbf{M}_{n} =
\begin{cases}
0, & n=n^*,\ n^* \in \mathcal{S}_{\mathrm{train}}
\quad \text{(in-region)}, \\
0, & n=n^*,\ n^* \in
\mathcal{S}\setminus\mathcal{S}_{\mathrm{train}}
\quad \text{(cross-region)}, \\
1, & n\neq n^*.
\end{cases}
\end{equation}
where $n^*$ is randomly selected station index, $\mathcal{S}$ is all stations, $\mathcal{S}_{\text{train}}$ is training set (in-region scope), and $\setminus$ is set minus. This masking strategy is primarily used for station interpolation and extrapolation to overcome observational gaps in spatial scales, which aligns with the core objectives of in-region and cross-region station generation tasks. It can also represent scenarios where all data within a certain data block is missing.

\subsection{UniGIO}

The overall framework of UniGIO is illustrated in Fig. \ref{method_fig}. It consists of two main stages: observation-driven complementary initialization and temporal dependency modeling.

In the first stage, the Observation Mixer (OM) integrates available GIO records from station, variable, and regional perspectives, while the Event Aligner (EA) further compensates for temporal phase shifts among neighboring stations. This stage aims to propagate sparse observed evidence to missing entries and provide a more complete and physically meaningful initialization. In the second stage, the Adaptive Temporal Mixer (ATM) models temporal dependencies in the initialized sequence with uncertainty-aware state transitions and pattern-decoupled MoE routing, enabling the model to adaptively handle drifting stable-extreme patterns under different observation reliabilities. Meanwhile, the Local Refiner (LR) reconstructs local temporal features from sampled latent representations through multi-receptive-field convolutions and selective gating, preserving local continuity weakened by MoE routing and CVAE sampling. These modules are organized within a simple CVAE framework, enabling unified and extreme event-controllable forecasting, imputation, and generation under arbitrary observation masks.

\begin{algorithm}[t]
\caption{UniGIO Pipeline}
\label{alg:unigio}
\begin{algorithmic}[1]
\REQUIRE Incomplete sequence $\mathbf{X}_M$, mask $\mathbf{M}$, metadata $\mathbf{I}$, ground truth $\mathbf{X}$
\ENSURE Completed sequence $\hat{\mathbf{X}}$

\STATE \textbf{Prior branch:}
\STATE $\mathbf{f}_{\text{C}} \leftarrow \operatorname{OM}(\mathbf{X}_M,\mathbf{M},\mathbf{I})$
\STATE $\mathbf{f}_{\text{T}} \leftarrow \operatorname{ATM}(\mathbf{f}_{\text{C}})$
\STATE $(\boldsymbol{\mu}_p,\boldsymbol{\sigma}_p) \leftarrow \operatorname{PriorMLP}(\mathbf{f}_{\text{T}})$
\STATE $p_\psi(\mathbf{z}|\mathbf{X}_M,\mathbf{M},\mathbf{I})
=\mathcal{N}(\boldsymbol{\mu}_p,\boldsymbol{\sigma}_p^2)$

\STATE \textbf{Posterior branch:}
\STATE $\mathbf{f}_{\text{S}} \leftarrow \operatorname{EA}(\mathbf{X},\mathbf{I})$
\STATE $\mathbf{f}_{\text{V}} \leftarrow \operatorname{Variable-wise-attention}(\mathbf{f}_{\text{S}})$
\STATE $\mathbf{f}_{\text{T}}' \leftarrow \operatorname{Bi-SSM}(\mathbf{f}_{\text{V}})$
\STATE $(\boldsymbol{\mu}_q,\boldsymbol{\sigma}_q) \leftarrow \operatorname{PostMLP}(\mathbf{f}_{\text{T}}')$
\STATE $q_\phi(\mathbf{z}|\mathbf{X})
=\mathcal{N}(\boldsymbol{\mu}_q,\boldsymbol{\sigma}_q^2)$
\STATE \textbf{Generation:}
\STATE Sample $\boldsymbol{\epsilon}\sim\mathcal{N}(\mathbf{0},\mathbf{I})$
\STATE Set pointwise sampling strength $\boldsymbol{\alpha}\in\mathbb{R}^{N\times T\times C}$
\STATE $\mathbf{z}\leftarrow \boldsymbol{\mu}_p+\boldsymbol{\alpha}\odot\boldsymbol{\sigma}_p\odot\boldsymbol{\epsilon}$
\STATE Adjust $\boldsymbol{\alpha}$ on target positions to control the intensity of extreme event generation
\STATE Decode $\hat{\mathbf{X}}\leftarrow p_\theta(\operatorname{LR}(\mathbf{z}), \mathbf{f}_{\text{T}})$
\end{algorithmic}
\end{algorithm}

Unlike existing time series methods that directly model incomplete sequences, UniGIO first recovers station and region level complementary representations before temporal modeling. This decoupled design reduces the burden of simultaneously inferring missing states and learning temporal evolution, allowing ATM to focus on a more complete and observation-constrained sequence. As a result, UniGIO preserves observation evidence while enabling efficient recovery of complete weather sequences within a lightweight unified framework. The overall training and inference pipeline is summarized in Algorithm~\ref{alg:unigio}.

\begin{figure}[!t]
\centering
\includegraphics[width=\linewidth,height=0.9\linewidth]{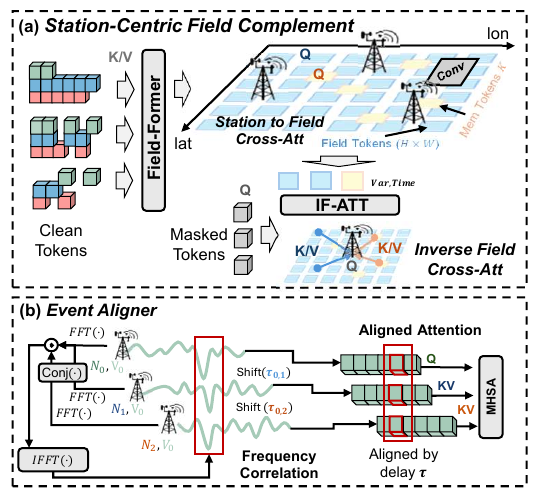}
\vspace{-17pt}
\caption{Detailed designs in Observation Mixer: (a) Station-Centric Field Complement to diffuse discrete station level observations into continuous local space with field and memory tokens; (b) Event Aligner corrects inter-station temporal delays via frequency-correlation alignment.}
\label{method_figb}
\end{figure}

\subsection{Observation Mixer and Event Aligner}

The Observation Mixer (OM) in Fig. \ref{method_fig} (b) is designed to initialize unobserved positions by exploiting the complementarity hidden in available GIO records. Since GIO is sparse and irregular, a missing value cannot be reliably recovered from broken temporal information alone. Instead, useful evidence may come from three levels: variables observed at the same station, neighboring stations at the same time, and the regional weather field formed by multiple stations. Therefore, OM performs self-complementarity, adjacent-complementarity, and field-complementarity in a progressive manner. The first two levels recover local station-wise information, while the field-level latent expands discrete station observations into a local continuous space where weather processes naturally evolve across multiple stations.

For self- and adjacent-complementarity, the OM applies self-attention $\operatorname{MHSA}_{\text{var}}(\cdot)$ over variables $\mathbf{f}_{\text{vars}}^t$ in each station feature $\mathbf{f}$, and aggregates neighboring stations via a distance-based graph attention $\operatorname{GAT}_{\text{sta}}(\cdot)$ with adjacency $\mathbf{A}$ calculated by $\operatorname{dist}(\cdot)$ with thresh $\tau$ at time step $t=1,\dots,T$. For these station-wise complementarity, the calculation is given by:

\begin{equation}
\begin{aligned}
& \mathbf{f} = \operatorname{MLP}(\mathbf{X}_M)+\operatorname{STE}(\mathbf{X}_M)
\in\mathbb{R}^{N\times T\times C \times d}, \\
& \mathbf{f}_{\text{vars}}^t = \mathbf{f}_{:,t,:,:}
\in\mathbb{R}^{N\times C\times d},
\quad t\in\{1,\dots,T\}, \\
& d_{ij}=\operatorname{dist}(\mathbf{s}_i,\mathbf{s}_j),\\
& \mathbf{A}_{ij}=
\begin{cases}
\exp(-d_{ij}/\tau), & d_{ij}\le \tau,\\
0, & d_{ij}>\tau,
\end{cases} \\
& \mathbf{f}_{\text{self}}^t
=\operatorname{MHSA}_{\text{var}}\big(\mathbf{f}_{\text{vars}}^t\big), \\
& \mathbf{f}_{\text{adj}}^t
=\operatorname{GAT}_{\text{sta}}\big(\mathbf{f}_{\text{self}}^t,\mathbf{A}\big),
\quad t=1,\dots,T .
\end{aligned}
\label{eq:station_mixer}
\end{equation}
where $\mathbf{f}_{\text{self}}$ are self variables complementary features, $\mathbf{f}_{\text{adj}}^t$ are adjacent station complement features, $d$ is latent size.

Then, beyond station-wise complementarity, our field-complementarity is detailed in Fig. \ref{method_figb} (a). Instead of projecting observations onto dense global grid at low resolution, OM constructs a small number of local scale field tokens $\mathbf{z}_{\text{field}}$ over gridded ($\operatorname{Grid}(\cdot)$) input station regions $\mathbf{G}_s$ with boundary $\phi$ and $\lambda$, $\delta$ is padding. Unlike heavy grids for global data, These tokens serve as lightweight regional queries that preserve station-centered dynamics in a continuous space with few $H \times W$ tokens. The field tokens are initialized with spatiotemporal encodings and interact with the adjacent-complemented station features through the cross-attention Field-former $\operatorname{Ff}(\cdot)$, which expands sparse GIO records into a continuous latent field. Meanwhile, $K$ learnable memory tokens $\mathbf{z}_{\text{mem}}$ are introduced to store global and event-level context shared across regional fields, enabling the model to maintain broader weather consistency beyond local station neighborhoods. Notably, both station tokens and field tokens share the same spatiotemporal encoding function $\operatorname{STE}(\cdot)$, which ensures that information propagation between discrete observations and continuous field latents remains spatially and temporally consistent throughout the process. For field-complementarity, calculation is given by:

\begin{equation}
\begin{aligned}
& \phi_{\min}=\min_i \text{lat}_i-\delta,\quad
\phi_{\max}=\max_i \text{lat}_i+\delta, \\
& \lambda_{\min}=\min_i \text{lon}_i-\delta,\quad
\lambda_{\max}=\max_i \text{lon}_i+\delta, \\
& \mathbf{G}_s
=\operatorname{Grid}\big(
[\phi_{\min},\phi_{\max}],
[\lambda_{\min},\lambda_{\max}],
H,W
\big), \\
& \mathbf{z}_{\text{field}}
=\operatorname{STE}(\mathbf{G}_s)
\in\mathbb{R}^{H\times W\times d}, \\
& \mathbf{z}_{\text{mem}}\in\mathbb{R}^{K\times d},
\quad \mathbf{z}_{\text{mem}}\ \text{is learnable}, \\
& [\mathbf{z}_{\text{field}}',\mathbf{z}_{\text{mem}}']
=
\operatorname{Ff}\big(
q=[\mathbf{z}_{\text{field}},\mathbf{z}_{\text{mem}}],
k/v=\{\mathbf{f}_{\text{adj}}^t\}_{t=1}^{T}
\big), \\
\end{aligned}
\label{eq:field_mixer}
\end{equation}

After the Field-former aggregates station evidence into field and memory tokens, CNNs are applied to enhance spatial continuity within the local field latent. Finally, OM uses masked tokens as queries in inverse cross-attention $\operatorname{ICAtt}(\cdot)$, projecting the continuous field representation back to the missing GIO positions and obtaining the complemented feature $\mathbf{f}_{\text{C}}$. The calculation is given by:

\begin{equation}
\begin{aligned}
& \mathbf{z}_{\text{ref}}
=\operatorname{CNNs}\big([\mathbf{z}_{\text{field}}',\mathbf{z}_{\text{mem}}']\big), \\
& \mathbf{f}_{\text{C}}
=
\operatorname{ICAtt}\big(
q=\mathbf{f}_{\text{mask}},
k/v=\mathbf{z}_{\text{ref}}
\big).
\end{aligned}
\label{eq:inverse_field_mixer}
\end{equation}

Although OM captures spatial and variable complementarity at the same time step, weather events usually reach different stations with temporal delays. Directly aggregating neighboring stations without considering such phase shifts may mix misaligned event states, especially for rapidly evolving local weather. To address this issue, we design the Event Aligner (EA) in Fig. \ref{method_figb} (b). EA estimates the pairwise delay matrix $\mathbf{D}\in \mathbb{R}^{N\times N}$ between stations using Fast Fourier Transform $\mathcal{F}(\cdot)$ cross-correlation with complex conjugation $\overline{\mathcal{F}(\cdot)}$. Then, neighboring station features are temporally shifted according to $\mathbf{D}$ before attention aggregation $\operatorname{MHA}(\cdot)$. This allows each station to attend to event-aligned neighboring features rather than raw features at the same timestamp.

\begin{equation}
\begin{aligned}
& \mathbf{D}_{i,j} =
\operatorname{argmax}
\Big(
\mathcal{F}^{-1}
\big(
\mathcal{F}(\mathbf{X}_i)
\odot
\overline{\mathcal{F}(\mathbf{X}_j)}
\big)
\Big), \\
& \mathbf{f}_a =
\operatorname{Shift}(\mathbf{f}_{\text{C}}, \mathbf{D})
\in \mathbb{R}^{N \times T \times N \times C \times d}, \\
& \mathbf{f}_{\text{ego}} =
\operatorname{Diag}(\mathbf{f}_a)
\in \mathbb{R}^{N \times T \times C \times d}, \quad
\mathbf{f}_{\text{adj}} =
\operatorname{OffDiag}(\mathbf{f}_a), \\
& \mathbf{f}_{\text{S}} =
\operatorname{MHA}
(\text{Q}:\mathbf{f}_{\text{ego}};
\text{K,V}:\mathbf{f}_{\text{adj}};
\text{Mask}:\mathbf{M}_s).
\end{aligned}
\label{eq:event_aligner}
\end{equation}

Here, $\odot$ is
dot product, $\operatorname{Diag}(\cdot)$ extracts each station's self-features from the aligned tensor, and $\operatorname{OffDiag}(\cdot)$ removes self-features to retain only neighboring features. Since reliable phase-delay estimation requires complete temporal sequences, EA is applied only in the posterior branch of the CVAE. In this way, EA uses complete target sequences during training to provide an aligned posterior correction, while the prior branch remains applicable to incomplete inputs during inference.

\subsection{Adaptive Temporal Mixer and Local Refiner}

\begin{figure}[!t]
\centering
\includegraphics[width=\linewidth,height=0.9\linewidth]{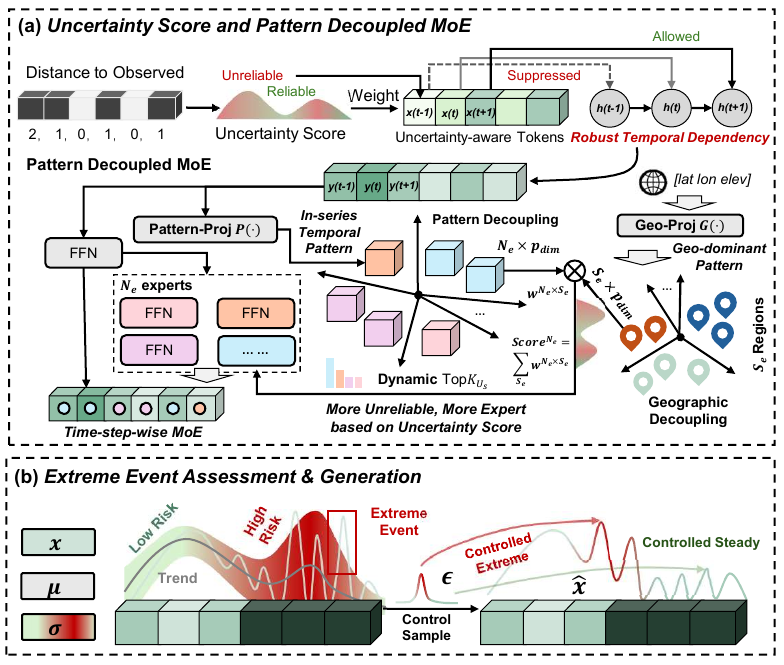}
\vspace{-17pt}
\caption {Details for extreme pattern capture, assessment, and generation.
(a) Uncertainty Score and Pattern Decoupled MoE suppress missing-induced state noise and decouple global similarity from local heterogeneity.
(b) Extreme event Assessment and Generation use CVAE variance and sampling-noise control to estimate and generate extremes.}
\label{method_figc}
\end{figure}

After complementary initialization, temporal modeling focuses on learning sequential dependencies between observed and unobserved parts. GIO sequences contain both smooth low-frequency evolution and abrupt local extremes, while missing observations introduce unreliable inputs. A single temporal mechanism may either overfit noisy missing regions or over-smooth short-term extremes. Therefore, we design the Adaptive Temporal Mixer (ATM) in Fig. \ref{method_fig} with a `robust SSM and adaptive MoE' strategy detailed in Fig. \ref{method_figc} (a). Specifically, ATM first uses a Bi-Mamba block to construct a denoised and robust temporal backbone from reliability-weighted inputs, and then employs a Pattern Decoupled MoE (PD-MoE) to adapt this backbone to heterogeneous stable-extreme weather patterns. This design separates trend extraction from pattern adaptation, allowing the model to preserve stable temporal dependencies while remaining sensitive to abrupt local changes.

For the state space model stage, ATM introduces an Uncertainty Score $\mathbf{U}_s$ to control how missing or low-reliability observations enter the state space. Different from treating all initialized tokens equally, $\mathbf{U}_s$ explicitly measures the reliability of each temporal position according to its distance to the nearest observed data. Positions farther from available observations are assigned lower reliability and are suppressed before state transition, preventing uncertain missing regions from dominating the temporal state. Given the observed temporal index set $\mathcal{T}_{obs}$, the uncertainty-aware input is formulated as:
\begin{equation}
\begin{aligned}
& d_t = \min_{t' \in \mathcal{T}_{obs}} |t - t'|,
\quad t \in \{1,\dots,T\}, \\
& \mathbf{U}_s^t =
\exp \Big(
-\operatorname{ReLU}
\big(
\operatorname{MLP}(d_t)
\big)
\Big), \\
& \tilde{\mathbf{f}}_{\text{C}}^t =
\mathbf{U}_s^t \odot \mathbf{f}_{\text{C}}^t .
\end{aligned}
\label{eq:uncertainty_score}
\end{equation}

Based on these reliability-weighted inputs $\tilde{\mathbf{f}}_{\text{C}}$, the Bi-Mamba block performs temporal modeling through compressed transitions on state space $\mathbf{h}$. Such compression acts as a denoising bottleneck for chaotic weather systems, filtering out severe perturbations that are weakly related to the dominant trend. In this way, the state space stage first obtains a stable low-frequency temporal representation $\mathbf{f}_{\text{T}}$, which serves as a robust dependency backbone for subsequent pattern adaptation:
\begin{equation}
\begin{aligned}
& \overrightarrow{\mathbf{h}}^t
=
\operatorname{SSM}_{f}
\big(
\tilde{\mathbf{f}}_{\text{C}}^1,\dots,\tilde{\mathbf{f}}_{\text{C}}^t
\big), \\
& \overleftarrow{\mathbf{h}}^t
=
\operatorname{SSM}_{b}
\big(
\tilde{\mathbf{f}}_{\text{C}}^T,\dots,\tilde{\mathbf{f}}_{\text{C}}^t
\big), \\
& \mathbf{f}_{\text{T}}^{t}
=
\operatorname{MLP}
\big(
[\overrightarrow{\mathbf{h}}^t;
\overleftarrow{\mathbf{h}}^t]
\big),
\in \mathbb{R}^{T\times d}.
\end{aligned}
\label{eq:bimamba_temporal}
\end{equation}

For the MoE stage, PD-MoE adapts the denoised temporal dependency to heterogeneous local weather patterns with $N_e$ independent experts. Although Bi-Mamba captures a stable low-frequency trend, local weather may still switch among smooth evolution, periodic fluctuations, and abrupt extreme mutations. Instead of using a single shared projection or an unconstrained MLP router, PD-MoE introduces a structured router composed of pattern clustering and region clustering. The pattern clustering function $\operatorname{P}(\mathbf{f}_{\text{T}}^t)$ maps temporal features into $N_e$ separated pattern subspaces with $p_{dim}$ size, grouping globally similar temporal states while separating consecutive mutations. The region clustering function $\operatorname{G}(\mathbf{I})$ maps station metadata and location features into $S_e$ geo-related subspaces with $p_{dim}$ size, allowing the routing process to distinguish regional responses under similar temporal patterns:
\begin{equation}
\begin{aligned}
& \mathbf{p}_t =
\operatorname{P}(\mathbf{f}_{\text{T}}^{t})
\in \mathbb{R}^{N_e\times p_{dim}}, \\
& \mathbf{g} =
\operatorname{G}(\mathbf{I})
\in \mathbb{R}^{S_e\times p_{dim}}, \\
& \mathbf{r}_t =
\operatorname{Softmax} \big(
\sum_{s \in S_e} \big(
\mathbf{p}_t \mathbf{g}^{\top}
\big)_s\big) \in \mathbb{R}^{N_e},
\end{aligned}
\label{eq:PD-MoE_router}
\end{equation}
where $\mathbf{p}_t$ denotes the temporal pattern affinity, $\mathbf{g}$ denotes the regional affinity, and $\mathbf{r}_t$ is the final routing score at time step $t$.

Moreover, MoE complements the sequential Mamba with a global pattern-level view. Unlike direct global attention, which may overemphasize large-magnitude jumps and lose stable temporal dependencies, the SSM-MoE structure separates the representation spaces of different patterns through Mamba and specialized experts, thereby balancing stable temporal dependency modeling and abrupt extreme event sensitivity. The number of activated experts $E_t$ is dynamically determined by relative $\mathbf{U}_s^t$: reliable tokens use fewer experts for sharper prediction, while high-uncertainty tokens aggregate more experts for flexible reconstruction. The expert $\mathcal{E}_t$ selection and feature $\mathbf{f}_{\text{T}}$ aggregation are formulated as:
\begin{equation}
\begin{aligned}
& E_t = \operatorname{Relative}(\mathbf{U}_s^t), \\
& \mathcal{E}_t =
\operatorname{TopK}
\big(
\mathbf{r}_t, K=E_t
\big), \\
& \mathbf{w}_{t,e} =
\frac{
\exp(\mathbf{r}_{t,e})
}{
\sum_{j\in\mathcal{E}_t}
\exp(\mathbf{r}_{t,j})
},
\quad e\in\mathcal{E}_t, \\
& \mathbf{f}_{\text{T}}^{t}
=
\sum_{e\in\mathcal{E}_t}
\mathbf{w}_{t,e}
\operatorname{MLP}_{e}
\big(
\mathbf{f}_{\text{T}}^{t}
\big)\\
& \mathbf{f}_{\text{T}}
=
[\mathbf{f}_{\text{T}}^{1},\mathbf{f}_{\text{T}}^{2},\dots,\mathbf{f}_{\text{T}}^{T}]
\end{aligned}
\label{eq:PD-MoE_expert}
\end{equation}

After ATM captures robust dependency and adaptive pattern transitions, the time-independent operations in MoE routing and CVAE sampling may weaken local temporal continuity. To address this issue, we design the Local Refiner (LR). LR applies convolutions with multiple receptive fields $\mathcal{R}=\{r_1,r_2,\dots,r_{N_r}\}$ to the sampled latent representation, reconstructing local temporal features at different scales. A selective MLP gate $\operatorname{SG}(\cdot)$ then adaptively selects among these components, allowing the model to preserve smooth local evolution while retaining necessary extreme variations.

For latent features $\mathbf{f}_{\text{T}} \in \mathbb{R}^{N \times T \times C \times d}$, LR proceeds as follows:

\begin{equation}
\begin{aligned}
& \mathbf{z} = \operatorname{Sample}(\operatorname{PriorMLP}(\mathbf{f}_{\text{T}}))
\in \mathbb{R}^{N \times T \times C \times d},\\
& \mathbf{z}_c = \operatorname{Concat}\left[\operatorname{Conv}(\mathbf{z}, r_{n_r})\right],
\quad {n_r} \in \{1, \dots, N_r\}, \\
& \mathbf{S} = \operatorname{Softmax}(\operatorname{SG}(\mathbf{f}_{\text{T}}))
\in \mathbb{R}^{N \times T \times C \times ({N_r}\times d/{N_r})}, \\
& \mathbf{\hat{X}} = \operatorname{MLP}
\left(\operatorname{Concat}\left[\mathbf{f}_{\text{T}};\mathbf{S}\odot\mathbf{z}_c\right]\right)
\in \mathbb{R}^{N \times T \times C}.
\end{aligned}
\end{equation}

In this way, ATM first obtains a denoised and robust low-frequency temporal backbone through compressed state space modeling, then uses uncertainty-aware expert routing to adapt to stable-extreme pattern shifts. LR further restores local temporal coherence after stochastic sampling and expert mixing, enabling UniGIO to model both smooth weather evolution and abrupt local changes.

\subsection{Loss Function}

Our total loss function is composed of the CVAE reconstruction loss, KL divergence loss, and orthonormal constraint regulations for pattern decoupling and spatial clustering:

Specifically, given the posterior distribution $q_\phi(\mathbf{z}|\mathbf{X})$, the prior distribution $p_\psi(\mathbf{z}|\mathbf{X}_M,\mathbf{M},\mathbf{I})$ and decoder $p_\theta(\mathbf{X}|\mathbf{z})$, the total objective is formulated as:
\begin{equation}
\begin{aligned}
\mathcal{L}
&= \underbrace{
-\mathbb{E}_{\mathbf{z} \sim q_\phi(\mathbf{z}|\mathbf{X})}
\left[
\log p_\theta(\mathbf{X}|\mathbf{z})
\right]
}_{\mathcal{L}_{rec}} \\
&\quad + \beta \underbrace{
D_{KL}
\left(
q_\phi(\mathbf{z}|\mathbf{X})
\|
p_\psi(\mathbf{z}|\mathbf{X}_M,\mathbf{M},\mathbf{I})
\right)
}_{\mathcal{L}_{KL}} \\
&\quad + \gamma \underbrace{
\left\|
\mathbf{P}^{\top}\mathbf{P}-\mathbf{I}
\right\|_F^2
}_{\mathcal{L}_{orth}} .
\end{aligned}
\label{eq:loss}
\end{equation}

where $\mathbf{P}$ denotes the learnable weights associated with the pattern decoupling function $\operatorname{P}(\cdot)$, the geo-related routing function $\operatorname{G}(\cdot)$, and the learnable memory tokens $\mathbf{z}_{\text{mem}}$. These basis functions are constrained to be orthonormal.

\section{Experiments}
\subsection{Benchmark and Setup}
\textbf{Dataset:} We curate and benchmark Global In-situ Weather Modeling tasks on the \textbf{GIO-U} dataset built from the Weather-5K \cite{han2024weather} (quality controlled on HadISD, ICOADS, et al.), which covers 5672 weather stations worldwide, recording hourly in-situ temperature, dew point, wind direction, wind rate, and sea-level pressure at each station from 2014 to 2024. We consider time windows of 72, 120, and 240 hours and use the 120-hour window as our main testbed.

\textbf{Mask Settings:} We highlight both \textbf{native} and \textbf{curated} masks. Lacking supervision, training on native missing data is inappropriate. The model was therefore trained on curated masks and \textbf{zero-shot tested} on native masks to assess under real-world incompleteness. Curated masks also support overall and task-specific evaluation.

\textbf{Baselines:}
We compare UniGIO with representative generative time series models and station-interaction-aware trajectory models. For general time series generation, we include TimeGAN \cite{yoon2019time}, TimeVAE \cite{desai2021timevae}, DiffWave \cite{kong2020diffwave}, SSSD \cite{lopezalcaraz2022diffusionbased}, and Diffusion-TS \cite{yuandiffusion}, covering GAN-, VAE-, and diffusion-based paradigms. These methods are widely used for time series generation, imputation, or distribution modeling, but they mainly focus on unconditional or weakly conditional sequence modeling and rarely consider the coupled station-variable dependencies inherent in GIO. Since few existing frameworks are directly designed for unified GIO forecasting, imputation, and generation, we further reimplement two recent trajectory generation methods, UniMTD \cite{yang2025unified} and SportsTraj \cite{xusports}, which naturally support interactions among multiple agents. We extend their interaction modeling from spatial trajectories to station and variable dependencies to align them with our task setting. In addition, we introduce a deterministic variant of UniGIO with the CVAE generative structure removed to verify the effectiveness of the generative paradigm.

\textbf{Evaluation Metrics:}
We evaluate model performance from three perspectives: accuracy, fidelity, and extreme event capture. For accuracy, we report MAE and MSE on masked regions to measure pointwise reconstruction or forecasting errors. For fidelity, we adopt FID \cite{yuandiffusion}, statistical error (STE), and Correlation Score (CroS) \cite{yuandiffusion} to assess distribution-level consistency between generated sequences and ground truth, including feature distribution, statistical variation, and cross-variable correlations. For extreme event capture, we improve SEDI \cite{han2024weather} into F1-SEDI at the 99.5th and 90.0th percentiles by combining SEDI-based recall with precision in an F1-score manner, since practical weather services require not only detecting extremes but also reducing false alarms. In task-specific comparisons, including forecasting and cross-region generation, we mainly report accuracy and F1-SEDI because practical GIO applications prioritize pointwise reliability and extreme event detection over distributional similarity. Since wind direction is a circular variable ranging from 0 to 360 degrees, F1-SEDI is not computed for wind direction.

For \textbf{Accuracy}, we adopt Mean Absolute Error (MAE) and Mean Squared Error (MSE) on mask area:

\begin{equation}
\begin{aligned}
\text{MAE} &= \frac{1}{\sum_{i} (1 - \mathbf{M}_i)} \sum_{i} (1 - \mathbf{M}_i)\, \left| \hat{\mathbf{X}}_i - \mathbf{X}_i \right| \\
\text{MSE} &= \frac{1}{\sum_{i} (1 - \mathbf{M}_i)} \sum_{i} (1 - \mathbf{M}_i)\, \left( \hat{\mathbf{X}}_i - \mathbf{X}_i \right)^2
\end{aligned}
\end{equation}
where $\sum_{i} (1 - M_i)$ is the total number of masked values. $i$ is element-wise index

For \textbf{Extreme Event Capture}, previous methods use Symmetric Extremal Dependence Index (SEDI) to represent the recall rate of extreme events. However, in practical applications, a large number of false alarms is also unacceptable. Therefore, we propose F1-SEDI, which evaluates both precision and recall in an F1-score manner:
\begin{equation}
\begin{aligned}
&\text{Precision} = \frac{1}{
\sum_i \mathbb{I}(\hat{\mathbf{X}}_i<Q^p_{\text{lower}}) +
\sum_i \mathbb{I}(\hat{\mathbf{X}}_i>Q^p_{\text{upper}})
} \\
&\quad \quad \Bigg(\sum_i \mathbb{I}(\hat{\mathbf{X}}_i<Q^p_{\text{lower}} \cap \mathbf{X}_i<Q^p_{\text{lower}}) + {} \\
&\quad \quad \sum_i \mathbb{I}(\hat{\mathbf{X}}_i>Q^p_{\text{upper}} \cap \mathbf{X}_i>Q^p_{\text{upper}})
\Bigg) \\
&\text{SEDI} = \frac{1}{
\sum_i \mathbb{I}(\mathbf{X}_i<Q^p_{\text{lower}}) +
\sum_i \mathbb{I}(\mathbf{X}_i>Q^p_{\text{upper}})
} \\
&\quad \quad \Bigg(\sum_i \mathbb{I}(\hat{\mathbf{X}}_i<Q^p_{\text{lower}} \cap \mathbf{X}_i<Q^p_{\text{lower}}) + {} \\
&\quad \quad \sum_i \mathbb{I}(\hat{\mathbf{X}}_i>Q^p_{\text{upper}} \cap \mathbf{X}_i>Q^p_{\text{upper}})
\Bigg) \\
&\text{F1-SEDI} = \frac{2 \times \text{SEDI} \times \text{Precision}}{\text{SEDI} + \text{Precision}}
\end{aligned}
\end{equation}
where $Q^p_{\text{lower}}$ and $Q^p_{\text{upper}}$ are the $p$th lower and upper percentiles, $\mathbb{I}$ is indicator function.

For \textbf{Fidelity}, we adopt Fr\'echet Inception Distance (FID) \cite{heusel2017gans}, statistical errors (STE), and Correlation Score (CroS) \cite{yoon2019time,yuandiffusion} between generated data distribution and ground truth data distribution:

\begin{equation}
\begin{aligned}
\text{FID}
&=
\left\|
\boldsymbol{\mu}_r-\boldsymbol{\mu}_f
\right\|_2^2
+
\operatorname{Tr}
\left(
\boldsymbol{\Sigma}_r
+
\boldsymbol{\Sigma}_f
-
2
\left(
\boldsymbol{\Sigma}_r
\boldsymbol{\Sigma}_f
\right)^{1/2}
\right)
\\
\text{STE} &=
\frac{
\sum_i \Bigg( (1-\mathbf{M}_i) \Bigg|
\sigma_r^w(\mathbf{X}_i) - \sigma_f^w(\hat{\mathbf{X}_i})
\Bigg| \Bigg)
}{
\sum_i (1-\mathbf{M}_i)
} \\
\text{CroS} &=
\frac{
\sum_i \Bigg( (1-\mathbf{M}_i)
\left|
\begin{aligned}
&\mathbb{E}_{N,T}\left[ \tilde{\hat{\mathbf{X}_i}} \odot \tilde{\hat{\mathbf{X}_i}}^\prime \right] \\
&\quad - \mathbb{E}_{N,T}\left[ \tilde{\mathbf{X}_i} \odot \tilde{\mathbf{X}_i}^\prime \right]
\end{aligned}
\right|
\Bigg)
}{
\sum_i (1-\mathbf{M}_i)
}
\end{aligned}
\end{equation}

For FID, $\boldsymbol{\mu}_r$, $\boldsymbol{\mu}_f$, $\boldsymbol{\Sigma}_r$, and $\boldsymbol{\Sigma}_f$ are the mean and covariance of ground truth and model output features. For STE, $\sigma_r^w(\mathbf{X})$ and $\sigma_f^w(\hat{\mathbf{X}})$ are the standard deviation of ground truth and model output in window $w$. For CroS, $\tilde{\hat{\mathbf{X}}}$ and $\tilde{\mathbf{X}}$ are normalized $\hat{\mathbf{X}}$ and $\mathbf{X}$; $\tilde{\hat{\mathbf{X}}}^\prime$ and $\tilde{\mathbf{X}}^\prime$ are lower-triangular variable pairs of normalized data; $\mathbb{E}_{N,T}[\cdot]$ is the expectation over station and time dimensions.

\subsection{Overall performance}
\subsubsection{Quantitative Results}
\begin{figure}[!t]
\centering
\includegraphics[width=\linewidth,height=0.84\linewidth]{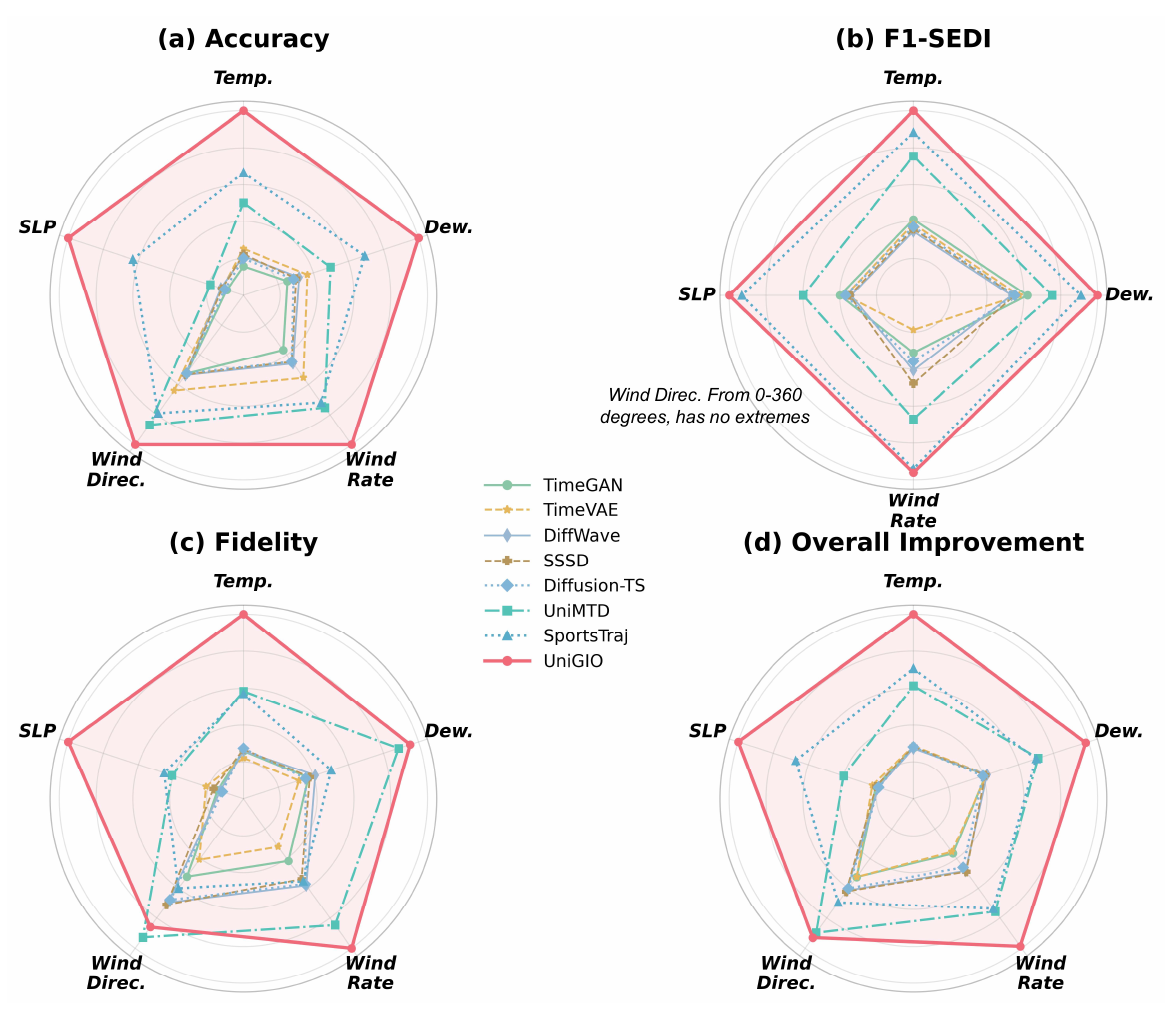}
\caption{UniGIO achieves SOTA performance on accuracy, extreme event capture, and fidelity across all time windows.}
\vspace{-10pt}
\label{overall_vis}
\end{figure}

Quantitative results on native masks and unified tasks with curated masks are visualized in Fig. \ref{overall_vis} and reported in Tab. \ref{tab:main_result_1}, Tab. \ref{tab:main_result_2}, and Tab. \ref{tab:main_result_3} in terms of accuracy, extreme event capture, and fidelity, respectively. The best results are highlighted in \textcolor{red}{red}, and the second-best results are highlighted in \textcolor{blue}{blue}. Our UniGIO achieves state-of-the-art performance over all baselines, leading by a large margin on the vast majority of 7 metrics and 5 variables on both curated and native masks.

For different mask settings, UniGIO outperforms the second-best baseline by an average of 12.1\% on accuracy, 3.5\% on F1-SEDI, and 11.1\% on fidelity over curated masks. And for the zero-shot transfer on native masks, UniGIO has achieved 10.8\%, 6.5\%, and 12.6\% advantage. The numerical performance is comparable with that on curated masks, which validates that our model can directly learn the intrinsic GIO patterns with real-world obscures from our synthetic mask designs.

For different methods, regrettably, both the deterministic method (UniGIO w/o CVAE) and time series generators fail on GIO data. The deterministic method is pulled to the mean value due to the MSE loss. Time series generators fall into two extremes: either mining only low-frequency and generating over-smoothed results, or focusing solely on high-frequency and producing spurious fluctuations on erroneous trends. Surprisingly, two trajectory methods survived and generated second-best results; the CVAE even achieved superiority on F1-SEDI compared with diffusion. The finding highlights the significance of both condition emphasized generative frameworks and GIO complementary.

We further compare representative frameworks under window lengths of 72h and 240h, with results reported in Tab. \ref{tab:main_result_window_Accuracy}, Tab. \ref{tab:main_result_window_F1-SEDI}, and Tab. \ref{tab:main_result_window_Fidelity}. UniGIO achieves the best performance across different temporal ranges. Under the 72h Curated/Native settings, it improves accuracy by 8.8\%/13.5\%, F1-SEDI by 3.6\%/3.6\%, and fidelity by 7.1\%/5.3\%. Under the 240h longer-window setting, the gains further increase to 11.1\%/10.0\% in accuracy, 5.1\%/6.0\% in F1-SEDI, and 21.4\%/16.4\% in fidelity, demonstrating its effectiveness under both short- and long-window scenarios, and UniGIO's advantage becomes more pronounced as the temporal window length increases. Moreover, as the window size increases, most metrics deteriorate due to longer temporal dependencies and higher uncertainty. CroS is an exception, as it measures inter-variable consistency and steadily improves with longer windows. This may be because longer windows provide richer temporal context for modeling variable relationships and reduce the sensitivity of this steady-state metric to short-window random fluctuations.

As shown in Table~\ref{tab:param_comparison}, the parameter comparison between UniGIO and the baseline methods demonstrates the lightness of our model design. Even among time series models, UniGIO remains compact, while UniGIO-M can be further compressed to less than 1M parameters. This further indicates that the improvement of UniGIO mainly stems from effective modeling designs rather than parameter scaling.

\begin{table*}[!t]
\centering
\caption{Quantitative Accuracy Comparison on Curated / Native Masks (120h) \scriptsize{$\dag$ means re-implemented} \vspace{-5pt}}
\label{tab:main_result_1}
\scriptsize
\renewcommand{\arraystretch}{0.8}
\begin{tabular*}{\textwidth}{@{\extracolsep{\fill}}@{}p{6em}@{}cccccccccc}
\toprule
\multirow{2}{*}{Baselines} &
\multicolumn{2}{c}{Temperature} &
\multicolumn{2}{c}{Dewpoint} &
\multicolumn{2}{c}{Wind Rate} &
\multicolumn{2}{c}{Wind Direc.} &
\multicolumn{2}{c}{Sea-Level Pressure} \\
\cmidrule(lr){2-3} \cmidrule(lr){4-5} \cmidrule(lr){6-7} \cmidrule(lr){8-9} \cmidrule(lr){10-11}
& MAE$\downarrow$ & MSE$\downarrow$ & MAE$\downarrow$ & MSE$\downarrow$ & MAE$\downarrow$ & MSE$\downarrow$ & MAE$\downarrow$ & MSE$\downarrow$ ($\times 10^3$) & MAE$\downarrow$ & MSE$\downarrow$ \\
\midrule
Deterministic & 4.54 / 4.32 & 25.5 / 26.8 & 3.37 / 3.54 & 16.7 / 17.5 & 2.10 / 2.16 & 6.19 / 5.98 & 92.8 / 97.4 & \textcolor{blue}{10.5} / 11.0 & 4.77 / 4.96 & 34.8 / 33.1 \\
TimeGAN & 5.51 / 10.3 & 37.8 / 76.5 & 3.91 / 6.98 & 22.4 / 41.5 & 2.51 / 4.33 & 8.31 / 14.7 & 103 / 147 & 12.5 / 17.9 & 5.77 / 8.69 & 49.8 / 78.5 \\
TimeVAE & 4.58 / 5.36 & 25.6 / 29.9 & 3.43 / 4.02 & 16.8 / 19.7 & 2.09 / 2.45 & 6.04 / 7.08 & 96.9 / 114 & 10.6 / 12.4 & 5.16 / 6.04 & 37.5 / 43.9 \\
Diffwave & 5.14 / 5.76 & 34.8 / 38.8 & 3.74 / 4.17 & 21.5 / 24.7 & 2.45 / 2.48 & 8.84 / 8.76 & 105 / 119 & 14.5 / 15.9 & 4.99 / 5.14 & 40.6 / 43.9 \\
SSSD & 4.98 / 5.39 & 33.1 / 37.9 & 3.88 / 4.25 & 23.9 / 27.0 & 2.37 / 2.69 & 8.65 / 9.52 & 107 / 111 & 15.4 / 16.1 & 5.00 / 5.78 & 41.6 / 48.1 \\
Diffusion-TS & 5.34 / 6.11 & 37.5 / 42.9 & 4.05 / 4.58 & 24.9 / 27.9 & 2.40 / 2.77 & 8.39 / 9.41 & 108 / 122 & 14.7 / 16.3 & 5.71 / 6.19 & 53.2 / 59.8 \\
UniMTD \dag & 2.69 / \textcolor{blue}{2.32} & 15.0 / 13.1 & 2.70 / \textcolor{blue}{2.30} & 16.6 / 13.7 & 1.68 / \textcolor{blue}{1.36} & 5.98 / \textcolor{blue}{4.82} & 81.1 / \textcolor{blue}{63.8} & 10.7 / \textcolor{blue}{7.47} & 3.69 / 3.03 & 33.8 / 26.3 \\
SportsTraj \dag & \textcolor{blue}{2.05} / 2.43 & \textcolor{blue}{7.75} / \textcolor{blue}{9.36} & \textcolor{blue}{1.97} / 2.31 & \textcolor{blue}{7.93} / \textcolor{blue}{9.70} & \textcolor{blue}{1.63} / 1.81 & \textcolor{blue}{4.92} / 5.32 & \textcolor{blue}{80.1} / 80.6 & 10.6 / 8.72 & \textcolor{blue}{1.44} / \textcolor{blue}{1.61} & \textcolor{blue}{4.37} / \textcolor{blue}{4.43} \\
\midrule
\textbf{UniGIO} & \textcolor{red}{1.45} / \textcolor{red}{1.51} & \textcolor{red}{5.35} / \textcolor{red}{6.01} & \textcolor{red}{1.50} / \textcolor{red}{1.45} & \textcolor{red}{5.99} / \textcolor{red}{6.06} & \textcolor{red}{1.27} / \textcolor{red}{1.22} & \textcolor{red}{3.63} / \textcolor{red}{3.61} & \textcolor{red}{64.6} / \textcolor{red}{60.1} & \textcolor{red}{8.45} / \textcolor{red}{7.04} & \textcolor{red}{1.02} / \textcolor{red}{0.91} & \textcolor{red}{3.67} / \textcolor{red}{1.83} \\
\bottomrule
\end{tabular*}
\vspace{-10pt}
\end{table*}

\begin{table*}[!t]
\centering
\caption{Quantitative F1-SEDI Comparison on Curated / Native Masks (120h) \scriptsize{$\dag$ means re-implemented} \vspace{-5pt}}
\label{tab:main_result_2}
\scriptsize
\renewcommand{\arraystretch}{0.8}
\begin{tabular*}{\textwidth}{@{\extracolsep{\fill}}@{}p{6em}@{}cccccccccc}
\toprule
\multirow{2}{*}{Baselines} &
\multicolumn{2}{c}{Temperature} &
\multicolumn{2}{c}{Dewpoint} &
\multicolumn{2}{c}{Wind Rate} &
\multicolumn{2}{c}{Wind Direc.} &
\multicolumn{2}{c}{Sea-Level Pressure} \\
\cmidrule(lr){2-3} \cmidrule(lr){4-5} \cmidrule(lr){6-7} \cmidrule(lr){8-9} \cmidrule(lr){10-11}
& 99.5th$\uparrow$ & 90.0th$\uparrow$& 99.5th$\uparrow$ & 90.0th$\uparrow$ & 99.5th$\uparrow$ & 90.0th$\uparrow$ & 99.5th$\uparrow$ & 90.0th$\uparrow$ & 99.5th$\uparrow$ & 90.0th$\uparrow$ \\
\midrule
Deterministic & 0.07 / 0.07 & 0.35 / 0.33  & 0.12 / 0.11 & 0.50 / 0.53 & 0.01 / 0.01 & 0.10 / 0.20 & $\backslash$ & $\backslash$ & 0.08 / 0.08 & 0.40 / 0.38 \\
TimeGAN & 0.10 / 0.19 & 0.46 / 0.48 & 0.21 / 0.28 & 0.58 / 0.60 & 0.03 / 0.06 & 0.19 / 0.26 & $\backslash$ & $\backslash$ & 0.19 / 0.19 & 0.50 / 0.51 \\
TimeVAE & 0.11 / 0.12 & 0.47 / 0.47 & 0.19 / 0.19 & 0.60 / 0.60 & 0.02 / 0.02 & 0.14 / 0.14 & $\backslash$ & $\backslash$ & 0.13 / 0.13 & 0.52 / 0.51 \\
Diffwave & 0.12 / 0.12 & 0.39 / 0.40 & 0.19 / 0.19 & 0.53 / 0.54 & 0.10 / 0.10 & 0.22 / 0.22 & $\backslash$ & $\backslash$ & 0.15 / 0.15 & 0.44 / 0.44 \\
SSSD & 0.12 / 0.13 & 0.40 / 0.42 & 0.20 / 0.21 & 0.53 / 0.55 & 0.12 / 0.12 & 0.25 / 0.26 & $\backslash$ & $\backslash$ & 0.15 / 0.16 & 0.44 / 0.46 \\
Diffusion-TS & 0.12 / 0.13 & 0.42 / 0.44 & 0.20 / 0.21 & 0.53 / 0.55 & 0.07 / 0.08 & 0.21 / 0.22 & $\backslash$ & $\backslash$ & 0.17 / 0.18 & 0.45 / 0.48 \\
UniMTD \dag & 0.34 / 0.44 & 0.66 / 0.72 & 0.28 / 0.43 & 0.60 / 0.67 & 0.12 / 0.21 & 0.32 / 0.44 & $\backslash$ & $\backslash$ & 0.35 / 0.47 & 0.59 / 0.65 \\
SportsTraj \dag & \textcolor{blue}{0.46} / \textcolor{blue}{0.50} & \textcolor{blue}{0.76} / \textcolor{blue}{0.78} & \textcolor{blue}{0.41} / \textcolor{blue}{0.47} & \textcolor{blue}{0.72} / \textcolor{blue}{0.76} & \textcolor{red}{0.26} / \textcolor{blue}{0.31} & \textcolor{blue}{0.38} / \textcolor{blue}{0.50} & $\backslash$ & $\backslash$ & \textcolor{blue}{0.73} / \textcolor{blue}{0.74} & \textcolor{blue}{0.85} / \textcolor{blue}{0.86} \\
\midrule
\textbf{UniGIO} & \textcolor{red}{0.55} / \textcolor{red}{0.61} & \textcolor{red}{0.81} / \textcolor{red}{0.83} & \textcolor{red}{0.46} / \textcolor{red}{0.56} & \textcolor{red}{0.75} / \textcolor{red}{0.80} & \textcolor{blue}{0.22} / \textcolor{red}{0.36} & \textcolor{red}{0.39} / \textcolor{red}{0.54} & $\backslash$ & $\backslash$ & \textcolor{red}{0.79} / \textcolor{red}{0.83} & \textcolor{red}{0.88} / \textcolor{red}{0.91} \\
\bottomrule
\end{tabular*}
\vspace{-10pt}
\end{table*}

\begin{table*}[!t]
\centering
\caption{Quantitative Fidelity Comparison on Curated / Native Masks (120h) \scriptsize{$\dag$ means re-implemented} \vspace{-5pt}}
\label{tab:main_result_3}
\scriptsize
\renewcommand{\arraystretch}{0.8}
\setlength{\tabcolsep}{1pt}
\begin{tabular*}{\textwidth}{@{\extracolsep{\fill}}lccccccccccc}
\toprule
\multirow{2}{*}{Baselines} &
\multicolumn{2}{c}{Temperature} &
\multicolumn{2}{c}{Dewpoint} &
\multicolumn{2}{c}{Wind Rate} &
\multicolumn{2}{c}{Wind Direc.} &
\multicolumn{2}{c}{Sea-Level Pressure} &
\multirow{2}{*}{CroS$\downarrow$} \\
\cmidrule(lr){2-3} \cmidrule(lr){4-5} \cmidrule(lr){6-7} \cmidrule(lr){8-9} \cmidrule(lr){10-11}
& FID$\downarrow$ & STE$\downarrow$ & FID$\downarrow$ & STE$\downarrow$ & FID$\downarrow$ & STE$\downarrow$ & FID$\downarrow$ & STE$\downarrow$ ($\times 10^3$) & FID$\downarrow$ & STE$\downarrow$ & \\
\midrule
Deterministic & 4.76 / 4.95 & 6.21 / 6.42 & 4.73 / 4.50 & 2.16 / 2.05 & 4.54 / 4.31 & 1.66 / 1.74 & 3.88 / 4.08 & 3.98 / 4.18 & 5.40 / 5.13 & 1.58 / 1.66 & 3.20 / 3.36 \\
TimeGAN & 1.22 / 0.95 & 5.53 / 5.38 & 1.11 / 0.89 & 2.32 / 2.29 & 1.08 / 0.78 & 1.50 / 1.52 & 0.95 / 0.72 & 3.80 / 3.80 & 2.97 / 2.51 & 2.53 / 2.51 & 3.10 / 3.17 \\
TimeVAE & 1.28 / 1.20 & 6.57 / 6.22 & 1.88 / 1.78 & 2.42 / 2.31 & 1.64 / 1.56 & 1.80 / 1.71 & 1.38 / 1.31 & 4.18 / 3.94 & 3.37 / 3.20 & 1.67 / 1.59 & 4.45 / 4.33 \\
Diffwave & 0.92 / 0.87 & 6.01 / 5.69 & 0.53 / 0.50 & 3.04 / 2.89 & 0.46 / 0.44 & 1.81 / 1.72 & 0.67 / 0.75 & 4.24 / 4.03 & 2.21 / 2.11 & 3.02 / 2.87 & 3.13 / 3.14 \\
SSSD & 0.82 / 0.78 & 6.05 / 5.75 & 0.51 / 0.48 & 3.77 / 3.58 & 0.48 / 0.36 & 2.01 / 1.91 & 0.64 / 0.71 & 4.65 / 4.42 & 1.28 / 1.20 & 2.54 / 2.41 & 3.16 / 3.11 \\
Diffusion-TS & 0.78 / 0.73 & 6.08 / 5.78 & 0.67 / 0.63 & 3.26 / 3.07 & \textcolor{blue}{0.43} / 0.40 & 1.54 / 1.46 & 0.65 / 0.79 & 3.90 / 3.71 & 1.72 / 1.63 & 3.67 / 3.44 & 3.64 / 3.59 \\
UniMTD \dag & \textcolor{blue}{0.50} / 0.27 & 3.34 / \textcolor{blue}{2.16} & \textcolor{blue}{0.30} / \textcolor{blue}{0.27} & \textcolor{blue}{1.64} / \textcolor{blue}{1.23} & 0.44 / \textcolor{blue}{0.35} & \textcolor{blue}{1.08} / \textcolor{blue}{0.78} & \textcolor{red}{0.52} / \textcolor{blue}{0.68} & \textcolor{blue}{2.73} / \textcolor{blue}{1.83} & 0.42 / 0.34 & 1.38 / \textcolor{blue}{0.96} & 2.44 / 1.42 \\
SportsTraj \dag & 0.53 / \textcolor{blue}{0.26} & \textcolor{blue}{3.27} / 2.37 & 0.77 / 0.39 & 2.26 / 1.69 & 0.87 / 0.47 & 1.33 / 1.00 & 1.44 / 0.78 & 3.08 / 2.07 & \textcolor{blue}{0.33} / \textcolor{blue}{0.25} & \textcolor{blue}{1.30} / 1.00 & \textcolor{blue}{1.52} / \textcolor{blue}{0.93} \\
\midrule
\textbf{UniGIO} & \textcolor{red}{0.17} / \textcolor{red}{0.13} & \textcolor{red}{2.48} / \textcolor{red}{1.67} & \textcolor{red}{0.25} / \textcolor{red}{0.21} & \textcolor{red}{1.45} / \textcolor{red}{1.02} & \textcolor{red}{0.28} / \textcolor{red}{0.19} & \textcolor{red}{1.07} / \textcolor{red}{0.68} &  \textcolor{blue}{0.60} / \textcolor{red}{0.52} & \textcolor{red}{2.53} / \textcolor{red}{1.67} & \textcolor{red}{0.11} / \textcolor{red}{0.09} & \textcolor{red}{0.79} / \textcolor{red}{0.51} & \textcolor{red}{1.46} / \textcolor{red}{0.81} \\
\bottomrule
\end{tabular*}
\vspace{-10pt}
\end{table*}

\begin{table*}[!t]
\centering
\caption{Quantitative Accuracy Comparison on Curated / Native Masks (72h/240h) \scriptsize{$\dag$ means re-implemented} \vspace{-5pt}}
\label{tab:main_result_window_Accuracy}
\scriptsize
\renewcommand{\arraystretch}{0.8}
\begin{tabular*}{\textwidth}{@{\extracolsep{\fill}}@{}p{6em}@{}c@{}cccccccccc}
\toprule
\multirow{2}{*}{Baselines} & \multirow{2}{*}{Window} &
\multicolumn{2}{c}{Temperature} &
\multicolumn{2}{c}{Dewpoint} &
\multicolumn{2}{c}{Wind Rate} &
\multicolumn{2}{c}{Wind Direc.} &
\multicolumn{2}{c}{Sea-Level Pressure} \\
\cmidrule(lr){3-4} \cmidrule(lr){5-6} \cmidrule(lr){7-8} \cmidrule(lr){9-10} \cmidrule(lr){11-12}
& & MAE$\downarrow$ & MSE$\downarrow$ & MAE$\downarrow$ & MSE$\downarrow$ & MAE$\downarrow$ & MSE$\downarrow$ & MAE$\downarrow$ & MSE$\downarrow$ ($\times 10^3$) & MAE$\downarrow$ & MSE$\downarrow$ \\
\midrule
\multirow{2}{*}{TimeGAN} & 72h & 5.24 / 9.42 & 33.2 / 63.4 & 3.33 / 5.71 & 16.3 / 28.7 & 2.49 / 4.20 & 8.05 / 13.8 & 97.0 / 136 & 11.4 / 15.9 & 4.55 / 6.49 & 31.1 / 45.0 \\
& 240h & 5.88 / 11.2 & 43.5 / 88.8 & 4.51 / 8.18 & 28.9 / 54.4 & 2.82 / 5.16 & 10.4 / 19.8 & 111 / 165 & 13.9 / 20.8 & 6.66 / 10.2 & 63.9 / 101 \\
\cmidrule(lr){1-12}
\multirow{2}{*}{TimeVAE} & 72h & 4.41 / 3.36 & 23.2 / 17.7 & 3.00 / 2.28 & 12.9 / 9.87 & 2.03 / 1.54 & 5.52 / \textcolor{blue}{4.20} & 93.0 / 70.9 & 10.1 / 7.66 & 4.24 / 3.23 & 25.8 / 19.6 \\
& 240h & 4.88 / 11.6 & 29.7 / 70.6 & 3.94 / 9.34 & 21.8 / 51.6 & 2.20 / 5.22 & 6.75 / 16.0 & 103 / 244 & 11.4 / 27.2 & 6.00 / 14.2 & 50.4 / 119 \\
\cmidrule(lr){1-12}
\multirow{2}{*}{UniMTD \dag} & 72h & 2.00 / 2.03 & 8.96 / 10.4 & 2.04 / \textcolor{blue}{1.99} & 9.94 / 10.7 & \textcolor{blue}{1.36} / \textcolor{blue}{1.28} & \textcolor{blue}{4.05} / 4.21 & \textcolor{blue}{66.2} / \textcolor{blue}{61.1} & \textcolor{red}{8.21} / \textcolor{blue}{7.08} & 2.65 / 2.35 & 19.8 / 17.2 \\
& 240h & 2.79 / \textcolor{blue}{2.66} & 16.4 / 17.1 & 2.97 / 2.71 & 19.3 / 18.5 & \textcolor{blue}{1.60} / \textcolor{blue}{1.47} & \textcolor{blue}{5.46} / \textcolor{blue}{5.48} & \textcolor{blue}{77.0} / \textcolor{blue}{69.7} & \textcolor{blue}{9.94} / \textcolor{blue}{8.53} & 4.54 / 3.64 & 47.1 / 36.3 \\
\cmidrule(lr){1-12}
\multirow{2}{*}{SportsTraj \dag} & 72h & \textcolor{blue}{1.65} / \textcolor{blue}{1.98} & \textcolor{blue}{5.18} / \textcolor{blue}{6.66} & \textcolor{blue}{1.80} / 2.08 & \textcolor{blue}{6.69} / \textcolor{blue}{8.19} & 1.54 / 1.69 & 4.34 / 4.62 & 75.8 / 77.0 & 9.96 / 8.45 & \textcolor{blue}{1.03} / \textcolor{blue}{1.14} & \textcolor{blue}{2.48} / \textcolor{blue}{2.50} \\
& 240h & \textcolor{blue}{2.37} / 2.82 & \textcolor{blue}{10.3} / \textcolor{blue}{12.0} & \textcolor{blue}{2.28} / \textcolor{blue}{2.68} & \textcolor{blue}{10.4} / \textcolor{blue}{12.2} & 1.77 / 1.99 & 5.71 / 6.16 & 86.5 / 87.2 & 11.9 / 9.64 & \textcolor{blue}{1.92} / \textcolor{blue}{2.15} & \textcolor{red}{7.26} / \textcolor{blue}{6.68} \\
\cmidrule(lr){1-12}
\multirow{2}{*}{UniGIO} & 72h & \textcolor{red}{1.26} / \textcolor{red}{1.37} & \textcolor{red}{4.15} / \textcolor{red}{5.56} & \textcolor{red}{1.32} / \textcolor{red}{1.33} & \textcolor{red}{4.80} / \textcolor{red}{5.93} & \textcolor{red}{1.22} / \textcolor{red}{1.14} & \textcolor{red}{3.34} / \textcolor{red}{3.47} & \textcolor{red}{63.1} / \textcolor{red}{56.1} & \textcolor{blue}{8.47} / \textcolor{red}{6.60} & \textcolor{red}{0.75} / \textcolor{red}{0.81} & \textcolor{red}{1.91} / \textcolor{red}{2.02} \\
& 240h & \textcolor{red}{1.72} / \textcolor{red}{1.71} & \textcolor{red}{7.34} / \textcolor{red}{7.63} & \textcolor{red}{1.76} / \textcolor{red}{1.65} & \textcolor{red}{8.16} / \textcolor{red}{7.80} & \textcolor{red}{1.36} / \textcolor{red}{1.28} & \textcolor{red}{4.12} / \textcolor{red}{3.32} & \textcolor{red}{68.9} / \textcolor{red}{62.3} & \textcolor{red}{9.15} / \textcolor{red}{8.52} & \textcolor{red}{1.47} / \textcolor{red}{1.27} & \textcolor{blue}{8.21} / \textcolor{red}{5.87} \\
\bottomrule
\end{tabular*}
\vspace{-10pt}
\end{table*}

\begin{table*}[!t]
\centering
\caption{Quantitative F1-SEDI Comparison on Curated / Native Masks (72h/240h) \scriptsize{$\dag$ means re-implemented} \vspace{-5pt}}
\label{tab:main_result_window_F1-SEDI}
\scriptsize
\renewcommand{\arraystretch}{0.8}
\begin{tabular*}{\textwidth}{@{\extracolsep{\fill}}@{}p{6em}@{}c@{}cccccccccc}
\toprule
\multirow{2}{*}{Baselines} & \multirow{2}{*}{Window} &
\multicolumn{2}{c}{Temperature} &
\multicolumn{2}{c}{Dewpoint} &
\multicolumn{2}{c}{Wind Rate} &
\multicolumn{2}{c}{Wind Direc.} &
\multicolumn{2}{c}{Sea-Level Pressure} \\
\cmidrule(lr){3-4} \cmidrule(lr){5-6} \cmidrule(lr){7-8} \cmidrule(lr){9-10} \cmidrule(lr){11-12}
& & 99.5th$\uparrow$ & 90.0th$\uparrow$& 99.5th$\uparrow$ & 90.0th$\uparrow$ & 99.5th$\uparrow$ & 90.0th$\uparrow$ & 99.5th$\uparrow$ & 90.0th$\uparrow$ & 99.5th$\uparrow$ & 90.0th$\uparrow$ \\
\midrule
\multirow{2}{*}{TimeGAN} & 72h & 0.15 / 0.23 & 0.49 / 0.51 & 0.28 / 0.34 & 0.65 / 0.66 & 0.04 / 0.08 & 0.24 / 0.29 & / & / & 0.35 / 0.35 & 0.62 / 0.63 \\
& 240h & 0.04 / 0.06 & 0.39 / 0.40 & 0.10 / 0.14 & 0.48 / 0.50 & 0.02 / 0.05 & 0.15 / 0.20 & / & / & 0.03 / 0.04 & 0.36 / 0.36 \\
\cmidrule(lr){1-12}
\multirow{2}{*}{TimeVAE} & 72h & 0.15 / 0.16 & 0.50 / 0.51 & 0.27 / 0.26 & 0.66 / 0.66 & 0.03 / 0.02 & 0.19 / 0.19 & / & / & 0.33 / 0.33 & 0.63 / 0.63 \\
& 240h & 0.03 / 0.03 & 0.40 / 0.39 & 0.05 / 0.06 & 0.50 / 0.50 & 0.02 / 0.01 & 0.09 / 0.10 & / & / & 0.01 / 0.01 & 0.34 / 0.34 \\
\cmidrule(lr){1-12}
\multirow{2}{*}{UniMTD \dag} & 72h & 0.48 / 0.49 & 0.75 / 0.75 & 0.37 / 0.45 & 0.68 / 0.71 & 0.17 / 0.24 & 0.37 / 0.46 & / & / & 0.51 / 0.55 & 0.70 / 0.72 \\
& 240h & 0.29 / 0.36 & 0.65 / 0.67 & 0.21 / 0.32 & 0.54 / 0.60 & 0.11 / 0.20 & 0.27 / 0.41 & / & / & 0.32 / 0.38 & 0.47 / 0.57 \\
\cmidrule(lr){1-12}
\multirow{2}{*}{SportsTraj \dag} & 72h & \textcolor{blue}{0.54} / \textcolor{blue}{0.56} & \textcolor{blue}{0.80} / \textcolor{blue}{0.82} & \textcolor{blue}{0.43} / \textcolor{blue}{0.48} & \textcolor{blue}{0.74} / \textcolor{blue}{0.77} & \textcolor{red}{0.23} / \textcolor{red}{0.31} & \textcolor{blue}{0.42} / \textcolor{blue}{0.52} & / & / & \textcolor{blue}{0.80} / \textcolor{blue}{0.80} & \textcolor{blue}{0.89} / \textcolor{blue}{0.89} \\
& 240h & \textcolor{blue}{0.40} / \textcolor{blue}{0.42} & \textcolor{blue}{0.72} / \textcolor{blue}{0.75} & \textcolor{blue}{0.30} / \textcolor{blue}{0.39} & \textcolor{blue}{0.68} / \textcolor{blue}{0.72} & \textcolor{red}{0.17} / \textcolor{red}{0.26} & \textcolor{blue}{0.31} / \textcolor{blue}{0.48} & / & / & \textcolor{blue}{0.63} / \textcolor{blue}{0.64} & \textcolor{blue}{0.80} / \textcolor{blue}{0.81} \\
\cmidrule(lr){1-12}
\multirow{2}{*}{UniGIO} & 72h & \textcolor{red}{0.60} / \textcolor{red}{0.63} & \textcolor{red}{0.83} / \textcolor{red}{0.85} & \textcolor{red}{0.51} / \textcolor{red}{0.58} & \textcolor{red}{0.78} / \textcolor{red}{0.81} & \textcolor{blue}{0.22} / \textcolor{blue}{0.28} & \textcolor{red}{0.43} / \textcolor{red}{0.55} & / & / & \textcolor{red}{0.85} / \textcolor{red}{0.83} & \textcolor{red}{0.92} / \textcolor{red}{0.91} \\
& 240h & \textcolor{red}{0.50} / \textcolor{red}{0.53} & \textcolor{red}{0.76} / \textcolor{red}{0.80} & \textcolor{red}{0.42} / \textcolor{red}{0.48} & \textcolor{red}{0.72} / \textcolor{red}{0.77} & \textcolor{blue}{0.15} / \textcolor{blue}{0.24} & \textcolor{red}{0.35} / \textcolor{red}{0.52} & / & / & \textcolor{red}{0.70} / \textcolor{red}{0.74} & \textcolor{red}{0.82} / \textcolor{red}{0.87} \\
\bottomrule
\end{tabular*}
\vspace{-10pt}
\end{table*}

\begin{table*}[!t]
\centering
\caption{Quantitative Fidelity Comparison on Curated / Native Masks (72h/240h) \scriptsize{$\dag$ means re-implemented} \vspace{-5pt}}
\label{tab:main_result_window_Fidelity}
\scriptsize
\renewcommand{\arraystretch}{0.8}
\setlength{\tabcolsep}{1pt}
\begin{tabular*}{\textwidth}{@{\extracolsep{\fill}}lcccccccccccc}
\toprule
\multirow{2}{*}{Baselines} & \multirow{2}{*}{Window} &
\multicolumn{2}{c}{Temperature} &
\multicolumn{2}{c}{Dewpoint} &
\multicolumn{2}{c}{Wind Rate} &
\multicolumn{2}{c}{Wind Direc.} &
\multicolumn{2}{c}{Sea-Level Pressure} &
\multirow{2}{*}{CroS$\downarrow$} \\
\cmidrule(lr){3-4} \cmidrule(lr){5-6} \cmidrule(lr){7-8} \cmidrule(lr){9-10} \cmidrule(lr){11-12}
& & FID$\downarrow$ & STE$\downarrow$ & FID$\downarrow$ & STE$\downarrow$ & FID$\downarrow$ & STE$\downarrow$ & FID$\downarrow$ & STE$\downarrow$ ($\times 10^3$) & FID$\downarrow$ & STE$\downarrow$ & \\
\midrule
\multirow{2}{*}{TimeGAN} & 72h & 0.47 / 0.39 & 5.67 / 5.54 & 0.42 / 0.34 & 2.16 / 2.15 & 0.42 / 0.32 & 1.47 / 1.49 & 0.90 / 0.91 & 3.78 / 3.78 & 1.18 / 1.06 & 2.03 / 2.00 & 3.42 / 3.50 \\
& 240h & 3.24 / 2.61 & 5.76 / 5.65 & 3.01 / 2.38 & 2.57 / 2.51 & 3.01 / 2.28 & 1.53 / 1.54 & 3.42 / 2.57 & 3.91 / 3.89 & 5.65 / 4.77 & 3.26 / 3.23 & 2.74 / 2.82 \\
\cmidrule(lr){1-13}
\multirow{2}{*}{TimeVAE} & 72h & 0.28 / 0.26 & 5.34 / 5.11 & 0.75 / 0.67 & 2.52 / 2.27 & 0.39 / 0.37 & 1.24 / 1.19 & 0.89 / 0.86 & 3.06 / 3.11 & 1.56 / 1.69 & 1.31 / 1.44 & 4.53 / 4.77 \\
& 240h & 2.74 / 2.43 & 6.72 / 6.74 & 3.04 / 3.37 & 2.48 / 2.42 & 4.11 / 3.80 & 1.85 / 1.80 & 4.06 / 3.93 & 4.26 / 4.13 & 6.15 / 4.90 & 1.68 / 1.61 & 4.24 / 4.25 \\
\cmidrule(lr){1-13}
\multirow{2}{*}{UniMTD $\dag$} & 72h & 0.19 / 0.22 & 3.07 / 2.03 & \textcolor{blue}{0.31} / \textcolor{blue}{0.17} & \textcolor{blue}{1.53} / \textcolor{blue}{1.16} & \textcolor{blue}{0.26} / \textcolor{blue}{0.30} & \textcolor{blue}{1.02} / \textcolor{blue}{0.66} & \textcolor{blue}{0.56} / \textcolor{blue}{0.67} & \textcolor{blue}{2.61} / \textcolor{blue}{1.59} & 0.15 / 0.12 & 1.26 / 0.87 & 2.56 / 1.52 \\
& 240h & \textcolor{blue}{1.82} / \textcolor{blue}{0.97} & \textcolor{blue}{3.85} / \textcolor{blue}{2.46} & \textcolor{blue}{1.07} / \textcolor{blue}{0.64} & \textcolor{blue}{1.74} / \textcolor{blue}{1.33} & \textcolor{blue}{1.66} / \textcolor{blue}{0.99} & \textcolor{blue}{1.13} / \textcolor{blue}{0.72} & \textcolor{blue}{2.03} / \textcolor{blue}{1.46} & \textcolor{blue}{2.86} / \textcolor{blue}{1.73} & \textcolor{blue}{1.23} / \textcolor{blue}{0.92} & \textcolor{blue}{1.46} / \textcolor{blue}{0.99} & 2.18 / 1.21 \\
\cmidrule(lr){1-13}
\multirow{2}{*}{SportsTraj $\dag$} & 72h & \textcolor{red}{0.12} / \textcolor{blue}{0.18} & \textcolor{blue}{2.59} / \textcolor{blue}{1.97} & 0.32 / 0.17 & 2.00 / 1.55 & 0.35 / 0.31 & 1.19 / 0.90 & 0.61 / 0.74 & 2.78 / 1.95 & \textcolor{blue}{0.12} / \textcolor{blue}{0.10} & \textcolor{blue}{0.84} / \textcolor{blue}{0.68} & \textcolor{blue}{1.70} / \textcolor{blue}{1.05} \\
& 240h & 1.90 / 0.99 & 3.90 / 2.83 & 2.28 / 1.27 & 2.83 / 2.05 & 2.87 / 1.67 & 1.55 / 1.15 & 4.30 / 2.48 & 3.48 / 2.26 & 1.52 / 1.02 & 2.16 / 1.52 & \textcolor{red}{1.24} / \textcolor{blue}{0.81} \\
\cmidrule(lr){1-13}
\multirow{2}{*}{UniGIO} & 72h & \textcolor{blue}{0.14} / \textcolor{red}{0.16} & \textcolor{red}{2.20} / \textcolor{red}{1.55} & \textcolor{red}{0.19} / \textcolor{red}{0.15} & \textcolor{red}{1.38} / \textcolor{red}{1.02} & \textcolor{red}{0.22} / \textcolor{red}{0.28} & \textcolor{red}{0.96} / \textcolor{red}{0.64} & \textcolor{red}{0.50} / \textcolor{red}{0.54} & \textcolor{red}{2.41} / \textcolor{red}{1.52} & \textcolor{red}{0.07} / \textcolor{red}{0.05} & \textcolor{red}{0.61} / \textcolor{red}{0.51} & \textcolor{red}{1.63} / \textcolor{red}{0.94} \\
& 240h & \textcolor{red}{0.30} / \textcolor{red}{0.32} & \textcolor{red}{2.86} / \textcolor{red}{1.78} & \textcolor{red}{0.43} / \textcolor{red}{0.45} & \textcolor{red}{1.53} / \textcolor{red}{1.17} & \textcolor{red}{0.50} / \textcolor{red}{0.44} & \textcolor{red}{1.10} / \textcolor{red}{0.69} & \textcolor{red}{1.02} / \textcolor{red}{0.97} & \textcolor{red}{2.67} / \textcolor{red}{1.62} & \textcolor{red}{0.17} / \textcolor{red}{0.12} & \textcolor{red}{1.05} / \textcolor{red}{0.65} & \textcolor{blue}{1.26} / \textcolor{red}{0.69} \\
\bottomrule
\end{tabular*}
\vspace{-10pt}
\end{table*}

\begin{table*}[!t]
\centering
\caption{Parameter Comparison of Unified and Task-specific GIO models. \vspace{-5pt}}
\label{tab:param_comparison}
\scriptsize
\renewcommand{\arraystretch}{0.9}
\setlength{\tabcolsep}{2.6pt}
\begin{tabular}{lccccccc|cc|cc|ccc}
\toprule
\textbf{Method}
& TimeGAN
& TimeVAE
& Diffwave
& SSSD
& Diffusion-TS
& UniMTD
& SportsTraj
& \textcolor{red}{UniGIO-M}
& \textcolor{red}{UniGIO}
& ESFM-S
& Chronos2
& WSSM
& iTransformer
& Corrformer \\
\midrule
\textbf{Params.}
& 3.4M
& 2.3M
& 4.1M
& 4.3M
& 4.9M
& 10.1M
& 7.3M
& \textcolor{red}{0.9M}
& \textcolor{red}{3.4M}
& 115M
& 120M
& 1.8M
& 0.9M
& 55.6M \\
\bottomrule
\end{tabular}
\vspace{-5pt}
\end{table*}

\subsubsection{Qualitative Results}
\label{Qualitative Results}
\begin{figure*}[!t]
\centering
\includegraphics[width=\textwidth,height=0.27\textwidth]{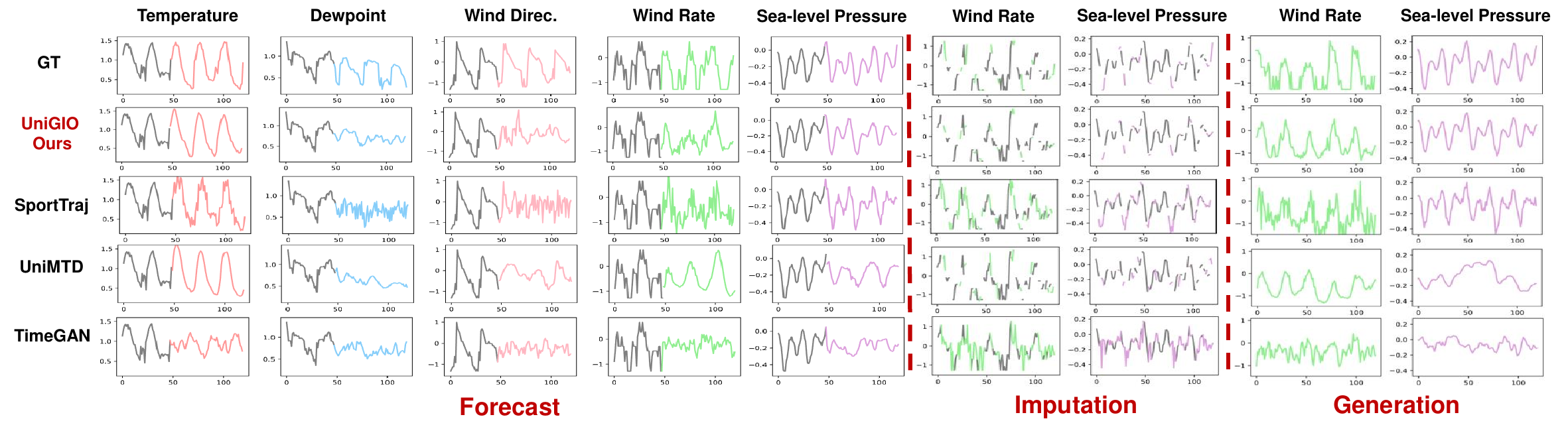}
\vspace{-17pt}
\caption{Due to space constraints, only wind rate and sea-level pressure are displayed for imputation and generation tasks}
\vspace{-10pt}
\label{vis_main}
\end{figure*}

In Fig. \ref{vis_main}, we qualitatively compared the forecasting, imputation, and generation results between UniGIO, SportsTraj, UniMTD, and TimeGAN. The results in row 2 illustrate that our UniGIO generates weather series with superior accuracy and fidelity, especially for temperature (\textcolor{red}{red}) and sea-level pressure (\textcolor{violet}{violet}). In row 3, SportsTraj generates a large number of spurious fluctuations, which are mainly attributed to the noise introduced by its sampling. Although this may help it simulate the random fluctuations of wind direction, it takes a toll on variables and tasks need inherent stability and continuity (e.g., temperature and imputation). Rows 4-5 show over-smoothing and over-fluctuation results, UniMTD can only generate low-frequency results, and this deficiency is not limited to high-frequency variables; it also leads to the loss of small-scale patterns in sea-level pressure. In contrast, TimeGAN merely fits high-frequency noise, with completely erroneous trends.

The qualitative evaluation results confirm the significant superiority of our method in quantitative results and also demonstrate heterogeneity among variables.

\begin{figure}[!t]
\centering
\includegraphics[width=\linewidth,height=0.75\linewidth]{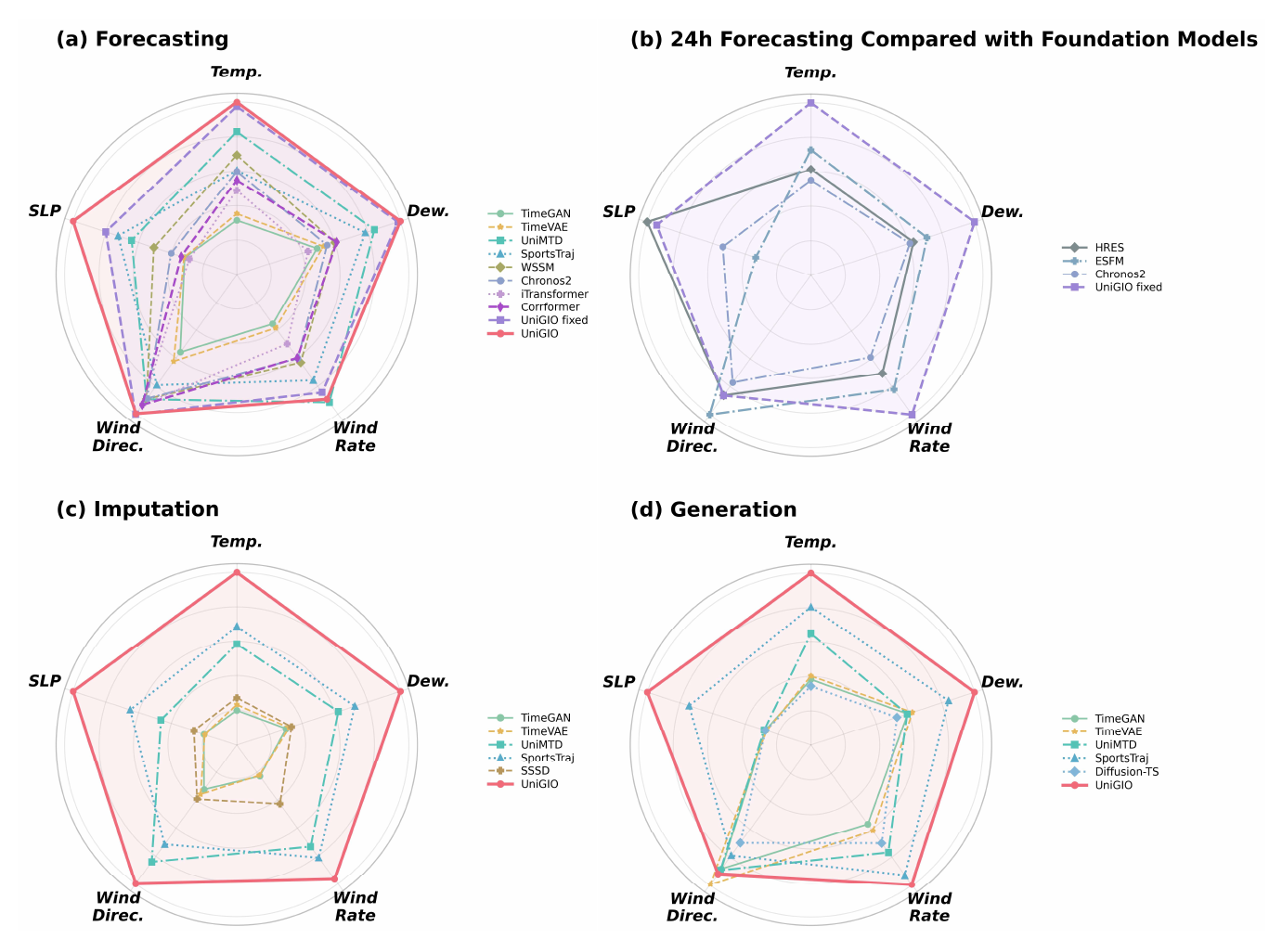}
\caption{UniGIO achieves SOTA performance on Forecasting, Imputation, and Generation tasks, even beating Foundation Models.}
\label{task_vis}
\end{figure}

\subsection{Task-specific Analysis}
\label{Task-specific Comparison}

\begin{table*}[!t]
\centering
\caption{Quantitative comparison under variable-length, fixed-length, and noisy forecasting settings.
\scriptsize{$\dag$ denotes re-implemented methods, and $\star$ denotes forecasting with an imputed look-back window.}
\vspace{-5pt}}
\label{tab:pred_result}
\scriptsize
\renewcommand{\arraystretch}{0.8}
\setlength{\tabcolsep}{0.5pt}
\begin{tabular*}{\textwidth}
{@{\extracolsep{\fill}}@{}p{6em}@{}cccccccccc}
\toprule
\multirow{2}{*}{Baselines} &
\multicolumn{2}{c}{Temperature} &
\multicolumn{2}{c}{Dewpoint} &
\multicolumn{2}{c}{Wind Rate} &
\multicolumn{2}{c}{Wind Direc.} &
\multicolumn{2}{c}{Sea-Level Pressure} \\
\cmidrule(lr){2-3}
\cmidrule(lr){4-5}
\cmidrule(lr){6-7}
\cmidrule(lr){8-9}
\cmidrule(lr){10-11}
&
MAE$\downarrow$/99.5th$\uparrow$ &
MSE$\downarrow$/90.0th$\uparrow$ &
MAE$\downarrow$/99.5th$\uparrow$ &
MSE$\downarrow$/90.0th$\uparrow$ &
MAE$\downarrow$/99.5th$\uparrow$ &
MSE$\downarrow$/90.0th$\uparrow$ &
MAE$\downarrow$/99.5th$\uparrow$ &
MSE$\downarrow$ ($\times 10^3$)/90.0th$\uparrow$ &
MAE$\downarrow$/99.5th$\uparrow$ &
MSE$\downarrow$/90.0th$\uparrow$ \\
\midrule

\multicolumn{11}{c}{\textbf{Variable-Length Forecasting}} \\
\midrule

TimeGAN
& 7.40 / 0.11
& 53.7 / 0.46
& 5.15 / 0.19
& 30.7 / 0.59
& 3.23 / 0.02
& 10.8 / 0.18
& 130.6 /
& 16.00 /
& 7.45 / 0.17
& 65.8 / 0.50 \\

TimeVAE
& 5.85 / 0.10
& 32.77 / 0.47
& 4.38 / 0.17
& 21.51 / 0.61
& 2.65 / 0.02
& 7.56 / 0.13
& 123.9 /
& 13.6 /
& 6.52 / 0.12
& 46.9 / 0.51 \\

UniMTD $\dag$
& \textcolor{blue}{2.35} / \textcolor{blue}{0.40}
& \textcolor{blue}{11.3} / \textcolor{blue}{0.70}
& \textcolor{blue}{2.27} / \textcolor{blue}{0.35}
& \textcolor{blue}{11.7} / \textcolor{blue}{0.65}
& \textcolor{blue}{1.59} / \textcolor{red}{0.22}
& \textcolor{blue}{5.29} / \textcolor{red}{0.35}
& \textcolor{blue}{76.3} /
& \textcolor{blue}{10.7} /
& 2.79 / 0.41
& 19.5 / 0.67 \\

SportsTraj $\dag$
& 3.32 / 0.30
& 17.2 / 0.55
& 2.61 / 0.32
& 12.1 / 0.64
& 1.94 / \textcolor{blue}{0.17}
& 6.14 / 0.30
& 88.9 /
& 11.7 /
& \textcolor{blue}{2.59} / \textcolor{blue}{0.42}
& \textcolor{blue}{12.9} / \textcolor{blue}{0.71} \\
\midrule
\textbf{UniGIO}
& \textcolor{red}{1.88} / \textcolor{red}{0.45}
& \textcolor{red}{7.85} / \textcolor{red}{0.75}
& \textcolor{red}{1.87} / \textcolor{red}{0.38}
& \textcolor{red}{8.23} / \textcolor{red}{0.71}
& \textcolor{red}{1.35} / 0.14
& \textcolor{red}{4.35} / \textcolor{blue}{0.33}
& \textcolor{red}{70.6} /
& \textcolor{red}{9.27} /
& \textcolor{red}{1.71} / \textcolor{red}{0.61}
& \textcolor{red}{8.28} / \textcolor{red}{0.78} \\

\midrule
\multicolumn{11}{c}{\textbf{Fixed-Length Forecasting}} \\
\midrule

WSSM
& \textcolor{blue}{2.65} / \textcolor{blue}{0.33}
& \textcolor{blue}{15.5} / \textcolor{blue}{0.62}
& \textcolor{blue}{2.89} / 0.21
& \textcolor{blue}{18.1} / \textcolor{blue}{0.53}
& 1.68 / \textcolor{blue}{0.09}
& 5.78 / \textcolor{blue}{0.20}
& \textcolor{blue}{75.9} /
& 10.80 /
& \textcolor{blue}{3.91} / \textcolor{blue}{0.36}
& \textcolor{blue}{28.2} / \textcolor{blue}{0.55} \\

Chronos2
& 2.83 / 0.23
& 17.5 / 0.58
& 2.95 / 0.19
& 22.3 / 0.50
& 1.55 / 0.05
& 5.02 / 0.16
& \textcolor{blue}{75.9} /
& 10.9 /
& 4.17 / 0.24
& 39.5 / 0.46 \\

iTransformer
& 3.29 / 0.15
& 21.98 / 0.52
& 3.54 / 0.10
& 25.91 / 0.45
& 1.70 / 0.01
& 6.17 / 0.16
& 77.83 /
& \textcolor{blue}{9.64} /
& 5.18 / 0.13
& 58.49 / 0.37 \\

Corrformer
& 3.00 / 0.17
& 17.72 / 0.56
& 3.16 / \textcolor{blue}{0.28}
& 20.88 / 0.51
& \textcolor{blue}{1.50} / 0.04
& \textcolor{blue}{4.44} / 0.12
& 76.56 /
& 9.74 /
& 4.30 / 0.15
& 40.91 / 0.40 \\
\midrule
\textbf{UniGIO}
& \textcolor{red}{1.96} / \textcolor{red}{0.45}
& \textcolor{red}{8.12} / \textcolor{red}{0.73}
& \textcolor{red}{1.92} / \textcolor{red}{0.38}
& \textcolor{red}{8.28} / \textcolor{red}{0.69}
& \textcolor{red}{1.43} / \textcolor{red}{0.12}
& \textcolor{red}{4.31} / \textcolor{red}{0.31}
& \textcolor{red}{70.4} /
& \textcolor{red}{9.24} /
& \textcolor{red}{2.23} / \textcolor{red}{0.49}
& \textcolor{red}{12.1} / \textcolor{red}{0.74} \\

\midrule
\multicolumn{11}{c}{\textbf{Forecasting with Incomplete Look-Back Window}} \\
\midrule

WSSM $\star$
& \textcolor{blue}{3.23} / \textcolor{blue}{0.13}
& \textcolor{blue}{20.1} / \textcolor{blue}{0.49}
& 3.26 / 0.11
& 21.84 / 0.47
& \textcolor{blue}{1.63} / 0.01
& 5.53 / 0.10
& \textcolor{blue}{74.7} /
& \textcolor{blue}{9.15} /
& 5.23 / 0.13
& 56.1 / 0.38 \\

Chronos2 $\star$
& 3.37 / 0.12
& 21.9 / 0.36
& \textcolor{blue}{2.06} / \textcolor{blue}{0.13}
& \textcolor{blue}{13.1} / 0.46
& 1.78 / \textcolor{blue}{0.02}
& 6.19 / \textcolor{blue}{0.11}
& 76.5 /
& 9.58 /
& \textcolor{blue}{4.75} / 0.10
& 54.5 / 0.32 \\

iTransformer $\star$
& 3.55 / 0.07
& 25.8 / 0.37
& 3.19 / 0.06
& 21.0 / 0.38
& 1.77 / 0.01
& 6.70 / 0.09
& 77.83 /
& 9.62 /
& 5.58 / 0.08
& 68.49 / 0.35 \\

Corrformer $\star$
& 3.32 / \textcolor{blue}{0.13}
& \textcolor{blue}{20.1} / 0.41
& 3.05 / 0.12
& 19.2 / \textcolor{blue}{0.48}
& 1.64 / 0.01
& \textcolor{blue}{5.36} / 0.09
& 75.34 /
& \textcolor{red}{9.08} /
& 4.83 / \textcolor{blue}{0.14}
& \textcolor{blue}{49.48} / \textcolor{blue}{0.39} \\
\midrule
\textbf{UniGIO}
& \textcolor{red}{2.06} / \textcolor{red}{0.40}
& \textcolor{red}{8.73} / \textcolor{red}{0.72}
& \textcolor{red}{2.02} / \textcolor{red}{0.33}
& \textcolor{red}{8.98} / \textcolor{red}{0.68}
& \textcolor{red}{1.48} / \textcolor{red}{0.12}
& \textcolor{red}{4.56} / \textcolor{red}{0.29}
& \textcolor{red}{73.2} /
& 9.64 /
& \textcolor{red}{2.31} / \textcolor{red}{0.48}
& \textcolor{red}{12.6} / \textcolor{red}{0.73} \\

\bottomrule
\end{tabular*}
\vspace{-10pt}
\end{table*}

\begin{table*}[!t]
\centering
\caption{Quantitative Comparison with Finetuned Observation and Time Series Foundation models on 24h Forecast Task\vspace{-5pt}}
\label{tab:pred_result_24h}
\scriptsize
\renewcommand{\arraystretch}{0.8}
\setlength{\tabcolsep}{0.5pt}  
\begin{tabular*}{\textwidth}{@{\extracolsep{\fill}}@{}p{6em}@{}cccccccccc}
\toprule
\multirow{2}{*}{Baselines} &
\multicolumn{2}{c}{Temperature} &
\multicolumn{2}{c}{Dewpoint} &
\multicolumn{2}{c}{Wind Rate} &
\multicolumn{2}{c}{Wind Direc.} &
\multicolumn{2}{c}{Sea-Level Pressure} \\
\cmidrule(lr){2-3} \cmidrule(lr){4-5} \cmidrule(lr){6-7} \cmidrule(lr){8-9} \cmidrule(lr){10-11}
& MAE$\downarrow$/99.5th$\uparrow$ & MSE$\downarrow$/90.0th$\uparrow$ & MAE$\downarrow$/99.5th$\uparrow$ & MSE$\downarrow$/90.0th$\uparrow$ & MAE$\downarrow$/99.5th$\uparrow$ & MSE$\downarrow$/90.0th$\uparrow$ & MAE$\downarrow$/99.5th$\uparrow$ & MSE$\downarrow$ ($\times 10^3$)/90.0th$\uparrow$ & MAE$\downarrow$/99.5th$\uparrow$ & MSE$\downarrow$/90.0th$\uparrow$ \\
\midrule
HRES & \textcolor{blue}{1.76} / - & \textcolor{blue}{7.39} / - & \textcolor{blue}{1.85} / - & \textcolor{blue}{7.94} / - & 1.48 / - & 4.53 / - & 63.8 /  & \textcolor{red}{7.15} /  & \textcolor{red}{0.86} / - & \textcolor{red}{2.68} / - \\
ESFM & \textcolor{blue}{1.76} / - & - / - & 1.86 / - & - / - & \textcolor{blue}{1.40} / - & - / - & \textcolor{red}{46.3} /  & - /  & 2.55 / - & - / - \\
Chronos2 & 2.05 / \textcolor{blue}{0.30} & 9.5 / \textcolor{blue}{0.58} & 2.06 / \textcolor{blue}{0.31} & 12.3 / \textcolor{blue}{0.65} & 1.47 / \textcolor{blue}{0.08} & \textcolor{blue}{4.51} / \textcolor{blue}{0.24} & 64.4 /  & 8.65 /  & 2.13 / \textcolor{blue}{0.52} & 11.3 / \textcolor{blue}{0.72} \\
\midrule
UniGIO fixed & \textcolor{red}{1.28} / \textcolor{red}{0.63} & \textcolor{red}{3.68} / \textcolor{red}{0.83} & \textcolor{red}{1.32} / \textcolor{red}{0.52} & \textcolor{red}{4.29} / \textcolor{red}{0.78} & \textcolor{red}{1.15} / \textcolor{red}{0.20} & \textcolor{red}{2.88} / \textcolor{red}{0.45} & \textcolor{blue}{58.6} /  & \textcolor{blue}{7.63} /  & \textcolor{blue}{1.05} / \textcolor{red}{0.76} & \textcolor{blue}{2.82} / \textcolor{red}{0.87} \\
\bottomrule
\end{tabular*}
\vspace{-10pt}
\end{table*}

\begin{table*}[!t]
\centering
\caption{Quantitative Comparison on Imputation Task \scriptsize{$\dag$ means re-implemented} \vspace{-5pt}}
\label{tab:Imp_result}
\scriptsize
\renewcommand{\arraystretch}{0.8}
\setlength{\tabcolsep}{0.5pt}  
\begin{tabular*}{\textwidth}{@{\extracolsep{\fill}}@{}p{6em}@{}cccccccccc}
\toprule
\multirow{2}{*}{Baselines} &
\multicolumn{2}{c}{Temperature} &
\multicolumn{2}{c}{Dewpoint} &
\multicolumn{2}{c}{Wind Rate} &
\multicolumn{2}{c}{Wind Direc.} &
\multicolumn{2}{c}{Sea-Level Pressure} \\
\cmidrule(lr){2-3} \cmidrule(lr){4-5} \cmidrule(lr){6-7} \cmidrule(lr){8-9} \cmidrule(lr){10-11}
& MAE$\downarrow$/99.5th$\uparrow$ & MSE$\downarrow$/90.0th$\uparrow$ & MAE$\downarrow$/99.5th$\uparrow$ & MSE$\downarrow$/90.0th$\uparrow$ & MAE$\downarrow$/99.5th$\uparrow$ & MSE$\downarrow$/90.0th$\uparrow$ & MAE$\downarrow$/99.5th$\uparrow$ & MSE$\downarrow$ ($\times 10^3$)/90.0th$\uparrow$ & MAE$\downarrow$/99.5th$\uparrow$ & MSE$\downarrow$/90.0th$\uparrow$ \\
\midrule
TimeGAN & 8.84 / 0.14 & 64.84 / 0.42 & 6.10 / 0.23 & 36.4 / 0.57 & 3.74 / 0.03 & 12.48 / 0.20 & 152 /  & 18.5 /  & 8.94 / 0.20 & 80.6 / 0.49 \\
TimeVAE & 6.77 / 0.16 & 37.7 / 0.47 & 5.14 / 0.22 & 25.3 / 0.60 & 3.10 / 0.02 & 8.87 / 0.14 & 145 /  & 15.9 /  & 7.71 / 0.14 & 56.3 / 0.52 \\
UniMTD \dag & 1.78 / 0.51 & 6.69 / 0.77 & \textcolor{blue}{1.61} / 0.49 & 6.20 / 0.75 & \textcolor{blue}{1.21} / 0.19 & \textcolor{blue}{3.13} / 0.42 & \textcolor{blue}{59.5} /  & \textcolor{blue}{7.01} /  & 1.89 / 0.62 & 9.23 / 0.78 \\
SportsTraj \dag & \textcolor{blue}{1.74} / \textcolor{blue}{0.60} & \textcolor{blue}{3.80} / \textcolor{blue}{0.84} & 1.65 / \textcolor{blue}{0.58} & \textcolor{blue}{3.88} / \textcolor{blue}{0.82} & 1.52 / \textcolor{red}{0.30} & 3.17 / \textcolor{blue}{0.49} & 73.9 /  & 7.87 /  & \textcolor{blue}{1.25} / \textcolor{blue}{0.82} & \textcolor{blue}{1.93} / \textcolor{blue}{0.91} \\
\midrule
SSSD & 5.49 / 0.22 & 35.21 / 0.51 & 4.51 / 0.26 & 28.23 / 0.59 & 2.52 / 0.17 & 8.85 / 0.27 & 116.82 /  & 16.46 /  & 5.67 / 0.31 & 47.54 / 0.57 \\
\midrule
UniGIO & \textcolor{red}{0.91} / \textcolor{red}{0.74} & \textcolor{red}{1.80} / \textcolor{red}{0.89} & \textcolor{red}{0.90} / \textcolor{red}{0.69} & \textcolor{red}{2.12} / \textcolor{red}{0.86} & \textcolor{red}{1.00} / \textcolor{blue}{0.26} & \textcolor{red}{2.09} / \textcolor{red}{0.53} & \textcolor{red}{49.0} /  & \textcolor{red}{6.01} /  & \textcolor{red}{0.55} / \textcolor{red}{0.92} & \textcolor{red}{0.61} / \textcolor{red}{0.95} \\
\bottomrule
\end{tabular*}
\vspace{-10pt}
\end{table*}

\begin{table*}[!t]
\centering
\caption{Quantitative Comparison on Generation Task \scriptsize{$\dag$ means re-implemented} \vspace{-5pt}}
\label{tab:Gen_result} 
\scriptsize
\renewcommand{\arraystretch}{0.8}
\setlength{\tabcolsep}{0.5pt}  
\begin{tabular*}{\textwidth}{@{\extracolsep{\fill}}@{}p{6em}@{}cccccccccc}
\toprule
\multirow{2}{*}{Baselines} &
\multicolumn{2}{c}{Temperature} &
\multicolumn{2}{c}{Dewpoint} &
\multicolumn{2}{c}{Wind Rate} &
\multicolumn{2}{c}{Wind Direc.} &
\multicolumn{2}{c}{Sea-Level Pressure} \\
\cmidrule(lr){2-3} \cmidrule(lr){4-5} \cmidrule(lr){6-7} \cmidrule(lr){8-9} \cmidrule(lr){10-11}
& MAE$\downarrow$/99.5th$\uparrow$ & MSE$\downarrow$/90.0th$\uparrow$ & MAE$\downarrow$/99.5th$\uparrow$ & MSE$\downarrow$/90.0th$\uparrow$ & MAE$\downarrow$/99.5th$\uparrow$ & MSE$\downarrow$/90.0th$\uparrow$ & MAE$\downarrow$/99.5th$\uparrow$ & MSE$\downarrow$ ($\times 10^3$)/90.0th$\uparrow$ & MAE$\downarrow$/99.5th$\uparrow$ & MSE$\downarrow$/90.0th$\uparrow$ \\
\midrule
TimeGAN & 3.81 / 0.12 & 23.9 / 0.47 & 2.80 / 0.20 & 15.2 / 0.59 & 1.87 / 0.03 & 6.15 / 0.20 & 77.7 /  & \textcolor{blue}{9.46} /  & 4.19 / 0.19 & 34.5 / 0.51 \\
TimeVAE & 3.46 / 0.11 & 19.3 / 0.47 & 2.57 / 0.18 & 12.5 / 0.60 & 1.59 / 0.02 & \textcolor{red}{4.63} / 0.15 & \textcolor{blue}{72.7} /  & \textcolor{red}{7.96} /  & 3.89 / 0.13 & 28.2 / 0.52 \\
UniMTD \dag & 2.60 / 0.33 & 14.15 / 0.67 & 2.83 / 0.23 & 17.2 / 0.57 & \textcolor{blue}{1.57} / 0.10 & 5.31 / 0.27 & 75.2 /  & 9.58 /  & 4.39 / 0.24 & 43.2 / 0.50 \\
SportsTraj \dag & \textcolor{blue}{2.10} / \textcolor{blue}{0.41} & \textcolor{blue}{9.06} / \textcolor{blue}{0.73} & \textcolor{blue}{2.04} / \textcolor{blue}{0.34} & \textcolor{blue}{9.35} / \textcolor{blue}{0.69} & 1.63 / \textcolor{red}{0.20} & 5.55 / \textcolor{blue}{0.35} & 80.8 /  & 11.6 /  & \textcolor{blue}{1.46} / \textcolor{blue}{0.71} & \textcolor{blue}{5.19} / \textcolor{blue}{0.84} \\
\midrule
Diffusion-TS & 4.23 / 0.13 & 33.06 / 0.42 & 3.06 / 0.19 & 20.44 / 0.54 & 1.94 / \textcolor{blue}{0.15} & 7.80 / 0.25 & 88.58 /  & 13.59 /  & 4.22 / 0.21 & 38.68 / 0.49 \\
\midrule
UniGIO & \textcolor{red}{1.61} / \textcolor{red}{0.49} & \textcolor{red}{6.06} / \textcolor{red}{0.77} & \textcolor{red}{1.67} / \textcolor{red}{0.40} & \textcolor{red}{7.01} / \textcolor{red}{0.73} & \textcolor{red}{1.48} / \textcolor{red}{0.20} & \textcolor{blue}{4.7} / \textcolor{red}{0.36} & \textcolor{red}{71.9} /  & 9.50 /  & \textcolor{red}{0.94} / \textcolor{red}{0.79} & \textcolor{red}{2.48} / \textcolor{red}{0.88} \\
\bottomrule
\end{tabular*}
\vspace{-10pt}
\end{table*}

To evaluate UniGIO as a versatile framework across different tasks on 120h time windows, in Fig. \ref{task_vis}, the accuracy and F1-SEDI performance of 7 tasks grouped into fixed/variable-window \textbf{Forecasting}, random\&period/resolution \textbf{Imputation}, and variable/in-region/cross-region \textbf{Generation} are presented from left to right. UniGIO achieved the optimal performance on 92\% of all 108 metrics. 

For forecasting, we obtained an advantage of 18.5\% and 8.13\% on accuracy and F1-SEDI. Notably, we significantly outperformed SOTA forecast methods, including WSSM \cite{yang2025wssm}, iTransformer \cite{liu2024itransformer}, Corrformer \cite{wu2023interpretable}, and Chronos2 \cite{ansari2024chronos}, in the 48h-72h fixed-window setting. More importantly, for forecasting with incomplete observations, our unified paradigm significantly outperforms existing approaches that first impute a complete look-back window before forecasting, yielding improvements of 24.5\% and 23.3\%, respectively. These results demonstrate that UniGIO is better suited for real-world forecasting under incomplete-observation GIO scenarios. For imputation and generation, our advantages reached 9.4\%, 6.0\% and 6.8\%, 4.4\%, respectively, also outperforming specialized imputation \cite{liu2019naomi,xu2023uncovering} and generation models \cite{yuandiffusion,yang2025unified,xusports}.

Further, UniGIO outperforms the fine-tuned global multi-source observation foundation model ESFM~\cite{ozdemir2026earth}, time series foundation model Chronos2 \cite{ansari2024chronos}, and operational NWP product HRES \cite{ecmwf2024hres} under the 24h forecast setting in ESFM, achieving an improvement of 14.7\%. These results verify that UniGIO, as a lightweight unified framework, is broadly effective across different sub-tasks. It can even surpass task-specific models and, despite model and data volume, outperform foundation models and operational NWP that rely on global multi-source observations or generic time series. Full tables are in Tab. \ref{tab:pred_result}, \ref{tab:pred_result_24h}, \ref{tab:Imp_result}, \ref{tab:Gen_result}.

\subsection{Spatial Generalization Analysis}
\label{Spatial Generalization Analysis}

\begin{table*}[!t]
\centering
\caption{Quantitative Evaluation on cross-region (South America / Africa) Generation \vspace{-5pt}}
\label{tab:main_result_cross}
\scriptsize
\renewcommand{\arraystretch}{0.8}
\begin{tabular*}{\textwidth}{@{\extracolsep{\fill}}@{}p{6em}@{}cccccccccc}
\toprule
\multirow{2}{*}{Settings} &
\multicolumn{2}{c}{Temperature} &
\multicolumn{2}{c}{Wind Rate} &
\multicolumn{2}{c}{Wind Direc.} &
\multicolumn{2}{c}{Temperature} &
\multicolumn{2}{c}{Wind Rate} \\
\cmidrule(lr){2-3} \cmidrule(lr){4-5} \cmidrule(lr){6-7} \cmidrule(lr){8-9} \cmidrule(lr){10-11}
& MAE$\downarrow$ & MSE$\downarrow$ & MAE$\downarrow$ & MSE$\downarrow$ & MAE$\downarrow$ & MSE$\downarrow$ ($\times 10^3$) & 99.5th$\uparrow$ & 90.0th$\uparrow$ & 99.5th$\uparrow$ & 90.0th$\uparrow$ \\
\midrule
World wide & 1.88 & 8.03 & 1.52 & 5.13 & 75.3 & 10.4 & 0.44 & 0.75 & 0.16 & 0.34 \\
\midrule
All stations & 3.14 / 2.22 & 20.5 / 11.3 & 1.81 / 1.62 & 6.91 / 5.77 & 80.9 / 84.9 & 11.7 / 12.2 & 0.13 / 0.29 & 0.46 / 0.64 & 0.08 / 0.12 & 0.20 / 0.23 \\
90\% stations & 2.28 / 2.07 & 15.9 / 9.67 & 1.83 / 1.52 & 7.48 / 4.77 & 74.0 / 82.1 & 10.2 / 11.4 & 0.22 / 0.27 & 0.60 / 0.68 & 0.08 / 0.08 & 0.25 / 0.24 \\
60\% stations & 2.06 / 1.84 & 8.76 / 7.99 & 1.68 / 1.36 & 6.11 / 3.78 & 73.4 / 80.9 & 10.1 / 11.1 & 0.32 / 0.30 & 0.64 / 0.71 & 0.14 / 0.10 & 0.27 / 0.25 \\
30\% stations & 2.04 / 1.70 & 8.20 / 7.17 & 1.65 / 1.35 & 5.76 / 3.05 & 72.7 / 79.8 & 10.4 / 10.7 & 0.36 / 0.33 & 0.70 / 0.72 & 0.13 / 0.16 & 0.28 / 0.27 \\
\bottomrule
\end{tabular*}
\vspace{-10pt}
\end{table*}

To evaluate spatial generalization, we conducted cross-domain generation on 120h time windows summarized in Tab. \ref{tab:main_result_cross}, generating African and South American stations using UniGIO trained on other stations. Focusing on temperature, wind direction, and wind rate—core variables in tropical coastal regions prone to storms and high temperatures.

Using in-region generation with fewer spatial gaps as a baseline `World wide', we randomly partitioned global stations into training and generation sets. For cross-region generation, first, the model was trained outside the target regions and tested on all stations in Africa and South America. Then, with 90\%, 60\%, or 30\% of stations randomly selected for generation and the remainder for fine-tuning.

Results in Tab. \ref{tab:main_result_cross} show South America is more challenging than Africa. Despite reduced performance under higher spatial incompleteness, the model maintains reasonable zero-shot accuracy, improving markedly when fine-tuned on only 10\% of stations. Fine-tuning on $>$50\% stations leads cross-region generation to outperform in-region generation on 35\% of 40 metrics, though extreme event capture still lags by 10\%. Overall, these results demonstrate the model's substantial spatial generalization capability.

\begin{figure}[!t]
\centering
\includegraphics[width=\linewidth,height=0.7\linewidth]{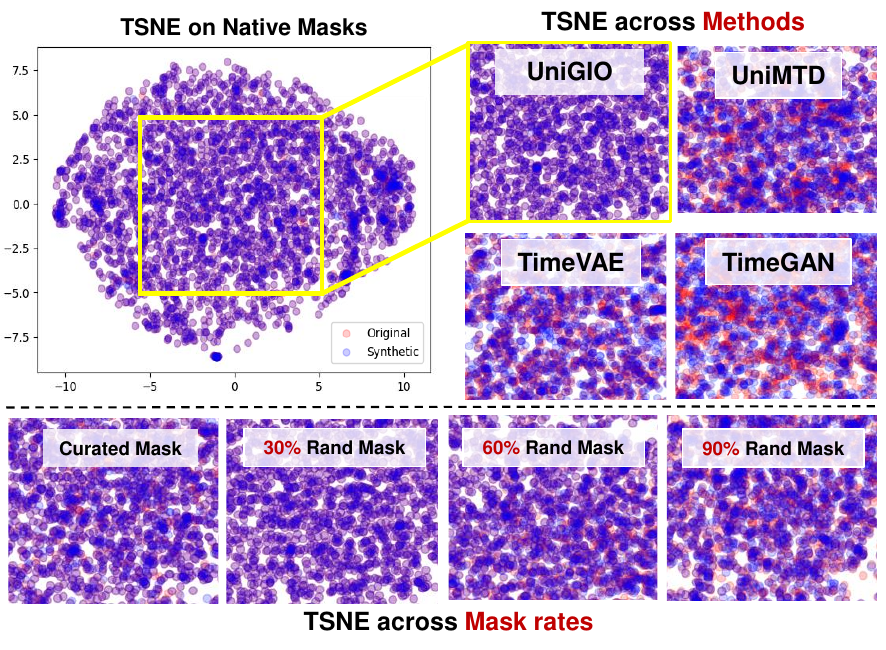}
\vspace{-17pt}
\caption{Distribution coverage of \textcolor{blue}{results} against \textcolor{red}{GT} across baselines and different mask settings. Less red area is better. TSNE plots look unclustered because GIO is largely continuous with no `category' between randomly selected samples for visualization.}
\label{tsne_vis}
\end{figure}

\subsection{Fidelity and Robustness Analysis}

In Fig. \ref{tsne_vis}, we present the distribution coverage of results on 120h time windows against ground truth across baselines and different mask settings under t-SNE reduction, comparing different baselines and mask settings. In this visualization, the exposed red regions indicate distributional discrepancies between the generated samples and the ground truth; therefore, a smaller visible red area suggests stronger distributional consistency and better coverage of the target data manifold.

As shown in the upper part, our method achieves more comprehensive coverage of the ground-truth distribution than the competing baselines. The generated samples produced by our method are more closely aligned with the ground-truth regions, while the baselines exhibit larger uncovered or mismatched areas. This observation indicates that our approach is better able to capture the underlying data distribution. Meanwhile, combined with the lower part, we observe that the native masks and the curated masks exhibit consistent alignment with the ground-truth distribution, indicating that our method is not sensitive to specific mask formats. These results confirm the performance advantages in the quantitative experiments.

For mask setting, despite mask type change or mask rates varying from 30\% (imputation-dominated) to 90\% (generation-dominated) with increasing red regions, our method maintains distribution consistency. Together with Exp \ref{Task-specific Comparison}, proving our robustness across varying conditions.

\begin{figure*}[!t]
\centering
\includegraphics[width=\textwidth,height=0.2\textwidth]{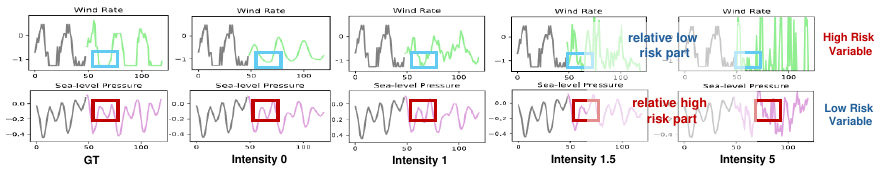}
\vspace{-15pt}
\caption{\textcolor{green}{Wind rate} with multiple extreme events and stable \textcolor{pink}{sea-level pressure} under different sampling noise intensities}
\label{ctrl_vis}
\end{figure*}

\subsection{Extreme Event Risk Analysis}

To verify whether the prior distribution variance of UniGIO serves as an indicator of the risk of extreme events, samples are generated under different noise intensities for two representative variables in Fig. \ref{ctrl_vis}: wind rate, which contains multiple extreme events, and sea-level pressure, which remains relatively stable.

It can be observed that when the intensity is $0$, both variables converge to a smooth mean trend. As the intensity increases, the former rapidly exhibits strong fluctuations, while the latter remains unchanged. When the intensity becomes excessively high, both variables fluctuate.

Notably, stable patterns exist even in high-risk variables, as highlighted by blue boxes, and vice versa. This observation suggests that the relationship between latent-space variance and extreme event risk is not uniformly distributed across all time steps, but instead reflects fine-grained, pointwise differences in uncertainty and event sensitivity.

These results indicate that we can, to a certain extent, quantify the risk of extreme events using the variance in the latent space. UniGIO also allows for pointwise control of extreme events as Fig. \ref{method_figc} (b), demonstrating not only interpretability in uncertainty modeling but also flexibility in controllable generation.

\begin{figure*}[!t]
\centering
\includegraphics[width=\linewidth,height=0.25\linewidth]{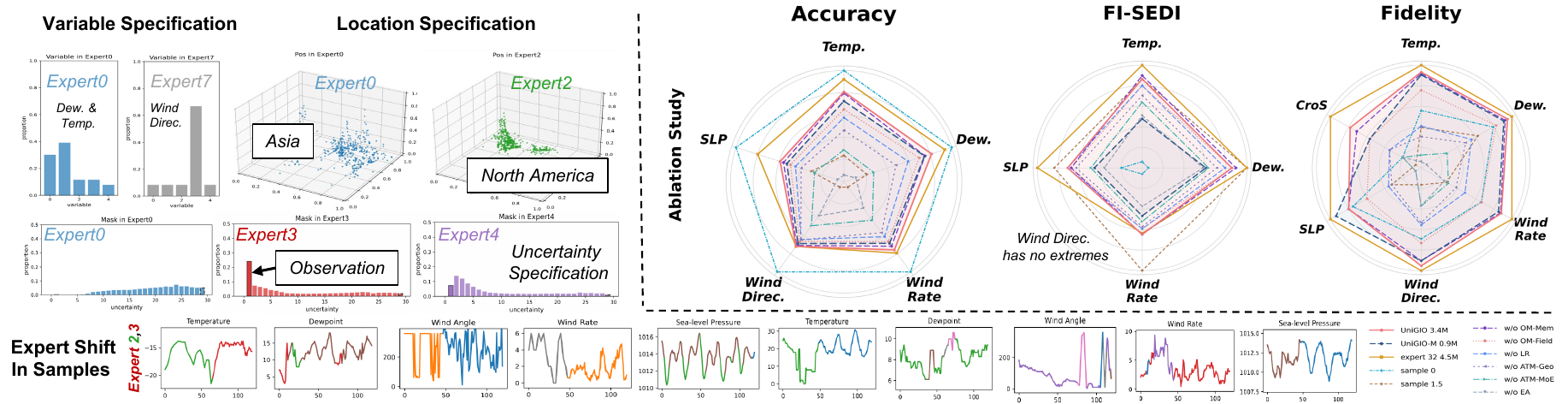}
\vspace{-15pt}
\caption{Ablation study results (right) and the representative expert specializations across variables, locations, uncertainties, and temporal patterns (left), different experts are denoted by colors}
\label{abl_vis}
\end{figure*}

\subsection{Model Structure Analysis}
\label{Ablation study}

\begin{table*}[!t]
\centering
\caption{Ablation Study on Accuracy and F1-SEDI\vspace{-5pt}}
\label{tab:abl_result_AF}
\scriptsize
\renewcommand{\arraystretch}{0.8}
\setlength{\tabcolsep}{0.5pt}  
\begin{tabular*}{\textwidth}{@{\extracolsep{\fill}}@{}p{7em}@{}cccccccccc} 
\toprule
\multirow{2}{*}{Baselines} &
\multicolumn{2}{c}{Temperature} &
\multicolumn{2}{c}{Dewpoint} &
\multicolumn{2}{c}{Wind Rate} &
\multicolumn{2}{c}{Wind Direc.} &
\multicolumn{2}{c}{Sea-Level Pressure} \\
\cmidrule(lr){2-3} \cmidrule(lr){4-5} \cmidrule(lr){6-7} \cmidrule(lr){8-9} \cmidrule(lr){10-11}
& MAE$\downarrow$/99.5th$\uparrow$ & MSE$\downarrow$/90.0th$\uparrow$ & MAE$\downarrow$/99.5th$\uparrow$ & MSE$\downarrow$/90.0th$\uparrow$ & MAE$\downarrow$/99.5th$\uparrow$ & MSE$\downarrow$/90.0th$\uparrow$ & MAE$\downarrow$/99.5th$\uparrow$ & MSE$\downarrow$ ($\times 10^3$)/90.0th$\uparrow$ & MAE$\downarrow$/99.5th$\uparrow$ & MSE$\downarrow$/90.0th$\uparrow$ \\
\midrule
UniGIO 3.4M & 1.45 / 0.55 & 5.35 / 0.81 & 1.50 / 0.46 & 5.99 / 0.75 & 1.27 / 0.22 & 3.63 / 0.39 & 64.6 /  & 8.45 /  & 1.02 / 0.79 & 3.67 / 0.88 \\
UniGIO-M 0.9M & 1.47 / 0.52 & 5.89 / 0.76 & 1.55 / 0.45 & 6.16 / 0.75 & 1.33 / 0.19 & 3.74 / 0.37 & 66.0 /  & 8.59 /  & 1.09 / 0.77 & 3.82 / 0.86 \\
expert 32 4.5M & 1.41 / 0.59 & 4.71 / 0.81 & 1.45 / 0.49 & 5.46 / 0.76 & 1.27 / 0.20 & 3.49 / 0.41 & 65.1 /  & 8.32 /  & 0.84 / 0.83 & 2.94 / 0.91 \\
sample 0 & 1.40 / 0.45 & 4.14 / 0.66 & 1.41 / 0.37 & 5.24 / 0.58 & 1.15 / 0.10 & 3.02 / 0.27 & 56.4 /  & 6.05 /  & 0.73 / 0.76 & 1.97 / 0.77 \\
sample 1.5 & 1.84 / 0.55 & 7.54 / 0.81 & 1.83 / 0.49 & 8.20 / 0.76 & 1.77 / 0.27 & 4.91 / 0.49 & 86.4 /  & 13.40 /  & 1.29 / 0.81 & 4.59 / 0.89 \\
\midrule
w/o OM-Mem & 1.46 / 0.56 & 5.39 / 0.81 & 1.54 / 0.47 & 6.18 / 0.75 & 1.29 / 0.22 & 3.74 / 0.39 & 65.1 /  & 8.49 /  & 1.05 / 0.79 & 3.69 / 0.87 \\
w/o OM-Field & 1.64 / 0.53 & 5.9 / 0.79 & 1.62 / 0.44 & 6.63 / 0.73 & 1.33 / 0.20 & 3.82 / 0.39 & 66.9 /  & 8.75 /  & 1.17 / 0.77 & 3.97 / 0.87 \\
w/o LR & 1.69 / 0.54 & 6.11 / 0.80 & 1.67 / 0.45 & 6.75 / 0.74 & 1.39 / 0.20 & 3.95 / 0.39 & 67.4 /  & 8.96 /  & 1.21 / 0.78 & 4.21 / 0.88 \\
w/o ATM-Geo & 1.71 / 0.53 & 6.53 / 0.78 & 1.73 / 0.44 & 6.97 / 0.74 & 1.48 / 0.19 & 4.07 / 0.38 & 67.2 /  & 8.95 /  & 1.28 / 0.77 & 4.26 / 0.87 \\
w/o ATM-MoE & 1.92 / 0.52 & 7.27 / 0.77 & 1.90 / 0.41 & 7.46 / 0.74 & 1.56 / 0.17 & 4.38 / 0.35 & 75.6 /  & 9.85 /  & 1.37 / 0.75 & 4.97 / 0.87 \\
w/o EA & 2.02 / 0.50 & 7.86 / 0.75 & 2.03 / 0.41 & 7.88 / 0.72 & 1.60 / 0.16 & 4.49 / 0.34 & 78.7 /  & 10.3 /  & 1.52 / 0.73 & 5.36 / 0.86 \\
\bottomrule
\end{tabular*}
\vspace{-10pt}
\end{table*}

\begin{table*}[!t]
\centering
\caption{Ablation Study on Fidelity \vspace{-5pt}}
\label{tab:abl_result_F}
\scriptsize
\renewcommand{\arraystretch}{0.8}
\setlength{\tabcolsep}{1pt}
\begin{tabular*}{\textwidth}{@{\extracolsep{\fill}}lccccccccccc}
\toprule
\multirow{2}{*}{Baselines} &
\multicolumn{2}{c}{Temperature} &
\multicolumn{2}{c}{Dewpoint} &
\multicolumn{2}{c}{Wind Rate} &
\multicolumn{2}{c}{Wind Direc.} &
\multicolumn{2}{c}{Sea-Level Pressure} &
\multirow{2}{*}{CroS$\downarrow$} \\
\cmidrule(lr){2-3} \cmidrule(lr){4-5} \cmidrule(lr){6-7} \cmidrule(lr){8-9} \cmidrule(lr){10-11}
& FID$\downarrow$ & STE$\downarrow$ & FID$\downarrow$ & STE$\downarrow$ & FID$\downarrow$ & STE$\downarrow$ & FID$\downarrow$ & STE$\downarrow$ ($\times 10^3$) & FID$\downarrow$ & STE$\downarrow$ & \\
\midrule
UniGIO 3.4M & 0.17 & 2.48 & 0.25 & 1.45 & 0.28 & 1.07 & 0.60 & 2.53 & 0.11 & 0.79 & 1.46 \\
UniGIO-M 0.9M & 0.19 & 2.47 & 0.29 & 1.46 & 0.30 & 1.09 & 0.64 & 2.63 & 0.05 & 0.71 & 1.49 \\
expert 32 4.5M & 0.14 & 2.45 & 0.21 & 1.43 & 0.23 & 0.97 & 0.55 & 2.46 & 0.02 & 0.68 & 1.43 \\
sample 0 & 0.29 & 2.77 & 0.41 & 1.48 & 0.54 & 1.10 & 0.94 & 2.83 & 0.19 & 0.72 & 1.54 \\
sample 1.5 & 0.27 & 3.09 & 0.45 & 1.87 & 0.66 & 1.57 & 1.25 & 4.11 & 0.26 & 1.15 & 1.56 \\
\midrule
w/o OM-Mem & 0.16 & 2.53 & 0.28 & 1.45 & 0.25 & 1.11 & 0.69 & 2.54 & 0.11 & 0.79 & 1.47 \\
w/o OM-Field & 0.24 & 2.58 & 0.45 & 1.51 & 0.39 & 1.24 & 0.97 & 2.63 & 0.19 & 0.95 & 1.49 \\
w/o LR & 0.30 & 2.99 & 0.54 & 1.68 & 0.54 & 1.33 & 1.21 & 2.81 & 0.27 & 1.14 & 1.52 \\
w/o ATM-Geo & 0.31 & 3.13 & 0.57 & 1.74 & 0.65 & 1.35 & 1.26 & 2.80 & 0.29 & 1.17 & 1.52 \\
w/o ATM-MoE & 0.39 & 3.21 & 0.68 & 1.91 & 0.79 & 1.34 & 1.45 & 3.02 & 0.36 & 1.24 & 1.54 \\
w/o EA & 0.43 & 3.21 & 0.75 & 2.17 & 0.85 & 1.32 & 1.42 & 3.08 & 0.39 & 1.28 & 1.54 \\
\bottomrule
\end{tabular*}
\vspace{-10pt}
\end{table*}

The right part of Fig. \ref{abl_vis} shows our ablation results on 120h time windows, corresponding to Tab. \ref{tab:abl_result_AF}, \ref{tab:abl_result_F}.

Regarding model capacity, performance increases with model size, increasing the number of experts from 16 (default) to 32 provides an effective way to improve performance. For varying sampling intensities, low intensity improves accuracy but degrades F1-SEDI, while high intensity has the opposite effect, both compromising fidelity. Despite using 73.5\% fewer parameters, UniGIO-M incurs only a 5\% performance compromise on average, while retaining largely comparable fidelity, demonstrating the strong parameter efficiency and compressibility of our framework.

Progressive structure ablation confirms the effectiveness of our design: The OM brings substantial improvements across all metrics, especially in accuracy and fidelity. This is consistent with the nature of weather events as field patterns rather than isolated pointwise series. MoEs consistently improve the performance across all variables, particularly in F1-SEDI. This validates that the ATM design is effective in capturing extreme events, rather than merely improving accuracy. LR enhances fidelity at the cost of smoothing extremes, and EA aids variables with distinct periodicity. The above results verify that the core design intentions of UniGIO are well fulfilled

The left and bottom parts of Fig. \ref{abl_vis} illustrate ATM expert specialization. Variable-wise, \textcolor{blue}{Expert0} favors temperature and dew point, while \textcolor{gray}{Expert7} handles wind direction. Location-wise, stations processed primarily by \textcolor{blue}{Expert0} cluster in Asia, whereas those handled by \textcolor{green}{Expert2} are concentrated in North America, Europe. Token-wise uncertainty shows \textcolor{blue}{Expert0} manages long-range high-uncertainty tokens, \textcolor{violet}{Expert4} short-range low-uncertainty tokens, and \textcolor{red}{Expert3} focuses on observed values. Expert shift in samples show reasonable pattern-expert allocation, despite token-wise routing, experts achieve segment-wise consistency due to robust long-term dependencies.


\section{Conclusion}
In this work, we aim to address global in-situ weather modeling under incomplete observations and establish a strong benchmark comprehensively evaluating accuracy, fidelity, and extreme event capture performance. Further, we propose UniGIO, a novel unified framework that allows in-situ weather modeling from incomplete GIO input through observation to missing generation. Essentially, we highlight (1) GIO complementarity; (2) Extreme patterns; (3) Lightweight to address the incompleteness and chaotic nature of GIO data and operate directly on interpolation-free distributed weather stations. We design an Observation Mixer and Event Aligner to capture the GIO complementarity at station and region level, and propose an Adaptive Temporal Mixer to divide-and-conquer stable evolution with chaotic drifting patterns. A Local Refiner is added to bridge the separated local patterns. Through extensive experiments and comprehensive evaluations, UniGIO consistently outperforms baselines across curated masks, native masks, and 7 forecasting, imputation, and generation tasks with an average 11\%, 12\%, and 5\% advantage on accuracy, fidelity, and extreme event capture. Establishing itself as a SOTA solution with significant potential for exploration in the climate foundation model.

\bibliographystyle{IEEEtran}
\bibliography{UniGIO}

\begin{IEEEbiography}
{Songru Yang}
received the B.S. degree from Beihang University, Beijing, China in 2023. He is pursuing the Ph.D. degree in the Department of Aerospace Intelligent Science and Technology, School of Astronautics, Beihang University. His research interests include machine learning, parttern recognition and AI4S.
\end{IEEEbiography}

\begin{IEEEbiography}[{\includegraphics[width=1in,height=1.25in,clip,keepaspectratio]{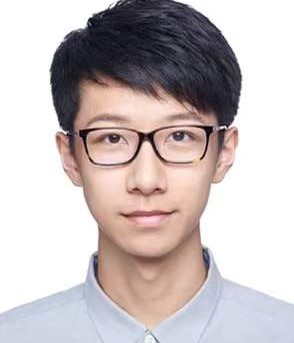}}]{Zili Liu} received the Ph.D. degree from the Image Processing Center, School of Astronautics, Beihang University, in 2025.

He is currently a Researcher at Shanghai Artificial Intelligence (AI) Laboratory, Shanghai, China. His research interests include deep learning, AI for meteorology, remote sensing image processing, and efficient deep learning.
\end{IEEEbiography}

\begin{IEEEbiography}[{\includegraphics[width=1in,height=1.25in,clip,keepaspectratio]{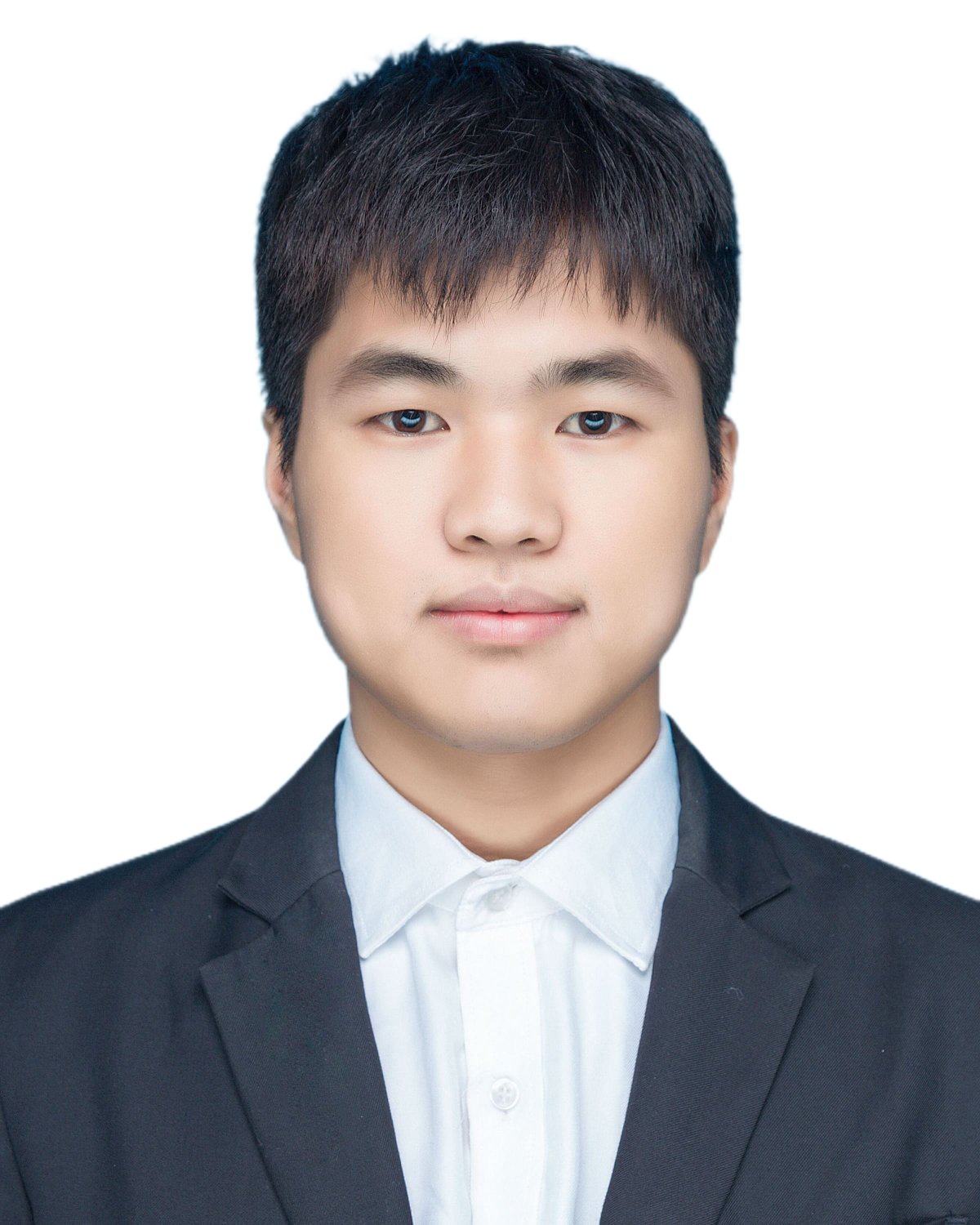}}]{Tao Han}received a B.E. degree in transportation equipment and control engineering and an M.S. degree in computer science and technology from Northwestern Polytechnical University, Xi’an, China, in 2019 and 2022. He is currently pursuing a Ph.D. degree in computer science and engineering at the Hong Kong University of Science and Technology. His research interests include computer vision, ai4science, and AIGC.
\end{IEEEbiography}

\begin{IEEEbiography}[{\includegraphics[width=1in,height=1.25in,clip,keepaspectratio]{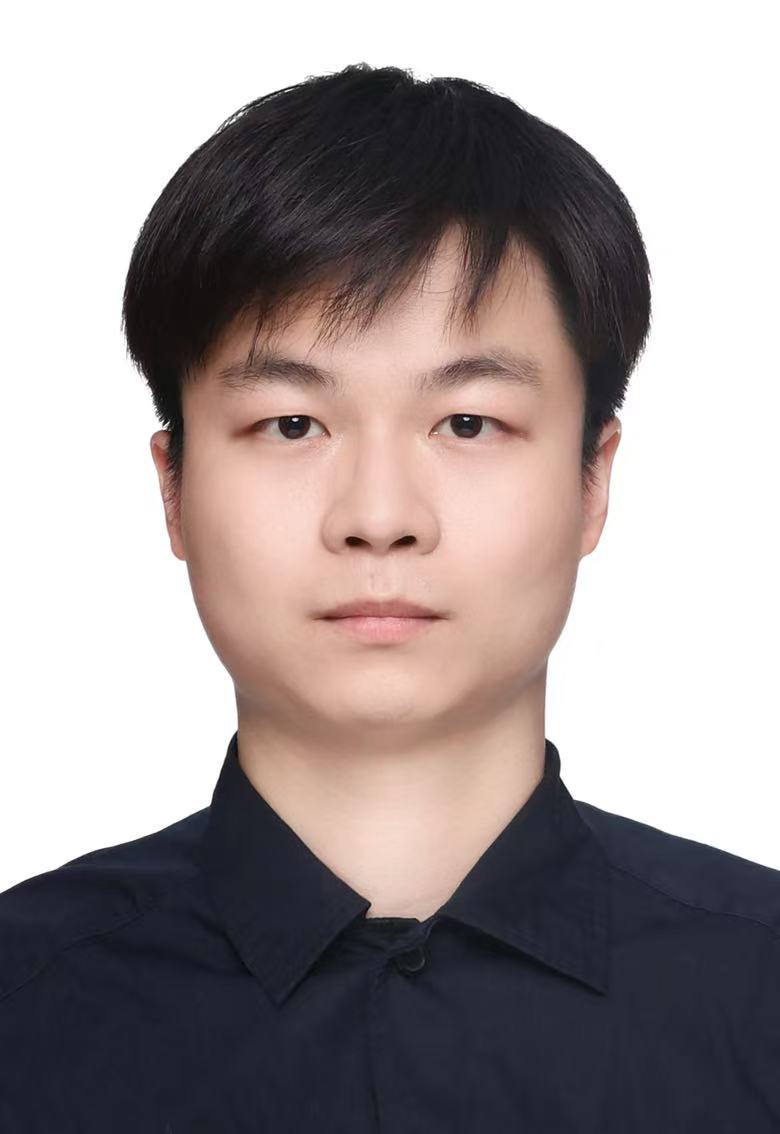}}]{Ben Fei} received the M.S. degree in the Department of Materials Science from Fudan University, Shanghai, China, in 2021. He received a Ph.D. degree in the School of Computer Science at Fudan University. He is currently a postdoctoral fellow at Multimedia Lab, Department of Information Engineering, The Chinese University of Hong Kong, Hong Kong SAR. His research interests include generative models, 3D computer vision and AI for Science. He is an IEEE Young Professional.
\end{IEEEbiography}

\begin{IEEEbiography}[{\includegraphics[width=1in,height=1.25in,clip,keepaspectratio]{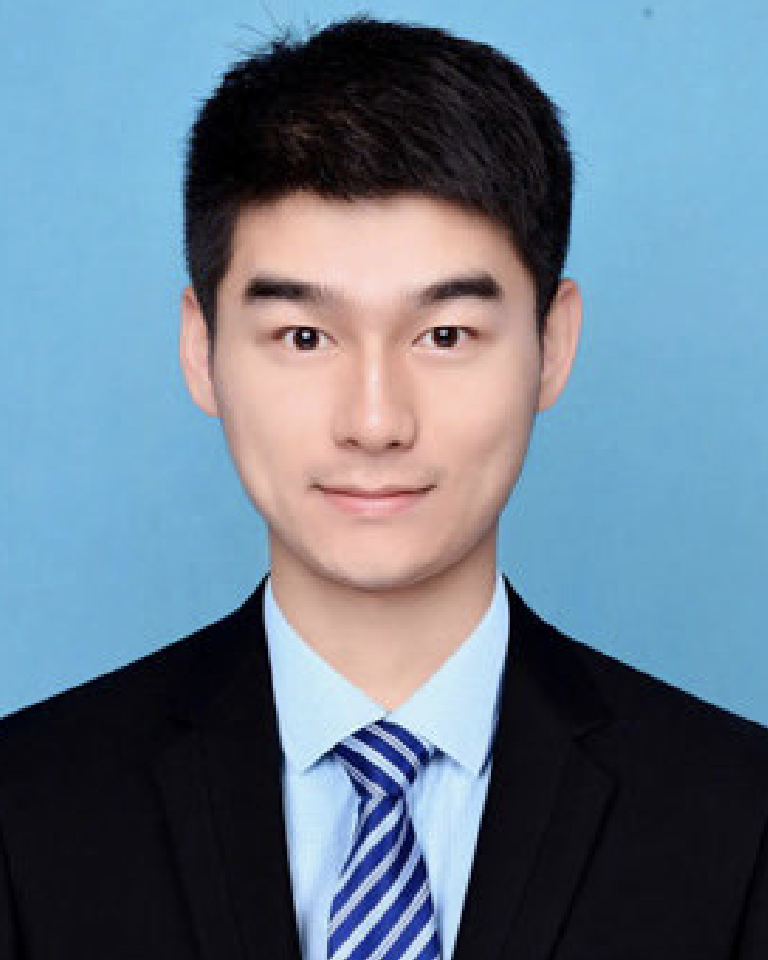}}]{Lei Bai}
received the Ph.D. degree from the University of New South Wales, Sydney, NSW, Australia, in 2021.

He was a Post-Doctoral Researcher with the University of Sydney, Camperdown, NSW, Australia.
He is currently a Research Scientist with Shanghai
Artificial Intelligence (AI) Laboratory, Shanghai,
China. He has authored or co-authored a set of
peer-reviewed papers in top AI conferences and
journals, such as Neural Information Processing
Systems (NeurIPS), Conference on Computer Vision
and Pattern Recognition (CVPR), International Joint Conference on Artificial Intelligence (IJCAI), Knowledge Discovery and Data Mining (KDD),
International Conference on Computer Vision (ICCV), International Conference on Ubiquitous Computing (Ubicomp), IEEE TRANSACTIONS ON PATTERN ANALYSIS AND MACHINE INTELLIGENCE (TPAMI), and IEEE TRANSACTIONS ON INTELLIGENT TRANSPORTATION SYSTEMS (TITS). His research interests include machine learning, spatial-temporal learning, and
their applications (e.g., Earth System Science and Smart City).

Dr. Bai is or was a Program Committee Member or Reviewer for IEEE TRANSACTIONS ON PATTERN ANALYSIS AND MACHINE INTELLIGENCE, NeurIPS, International Conference on Machine Learning (ICML), International Conference on Learning Representations (ICLR), CVPR, ICCV, Association for the Advancement of Artificial Intelligence (AAAI), IJCAI, KDD, European Conference on Computer Vision (ECCV), IEEE TRANSACTIONS ON IMAGE PROCESSING, IEEE TRANSACTIONS ON MULTIMEDIA, and ACM Transactions on Sensor Networks. He was a recipient of
the 2020 Google Ph.D. Fellowship, the 2020 UNSW Engineering Excellence Award, and the 2021 Dean’s Award for Outstanding Ph.D. Theses.
\end{IEEEbiography}

\begin{IEEEbiography}[{\includegraphics[width=1in,height=1.25in,clip,keepaspectratio]{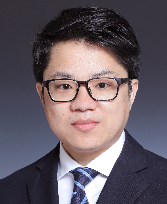}}]{Chang Liu} received the B.S. degree from Jilin University, Changchun, Jilin, China, in 2012, and the Ph.D. degree from the University of Chinese Academy of Sciences, Beijing, China, in 2022. He is currently a Post-Doctoral Researcher with the Department of Automation, School of Information Science and Technology, Tsinghua University, Beijing. He has published more than 50 papers in refereed conferences and journals. His research interests include computer vision and machine learning.
\end{IEEEbiography}

\begin{IEEEbiography}[{\includegraphics[width=1in,height=1.25in,clip,keepaspectratio]{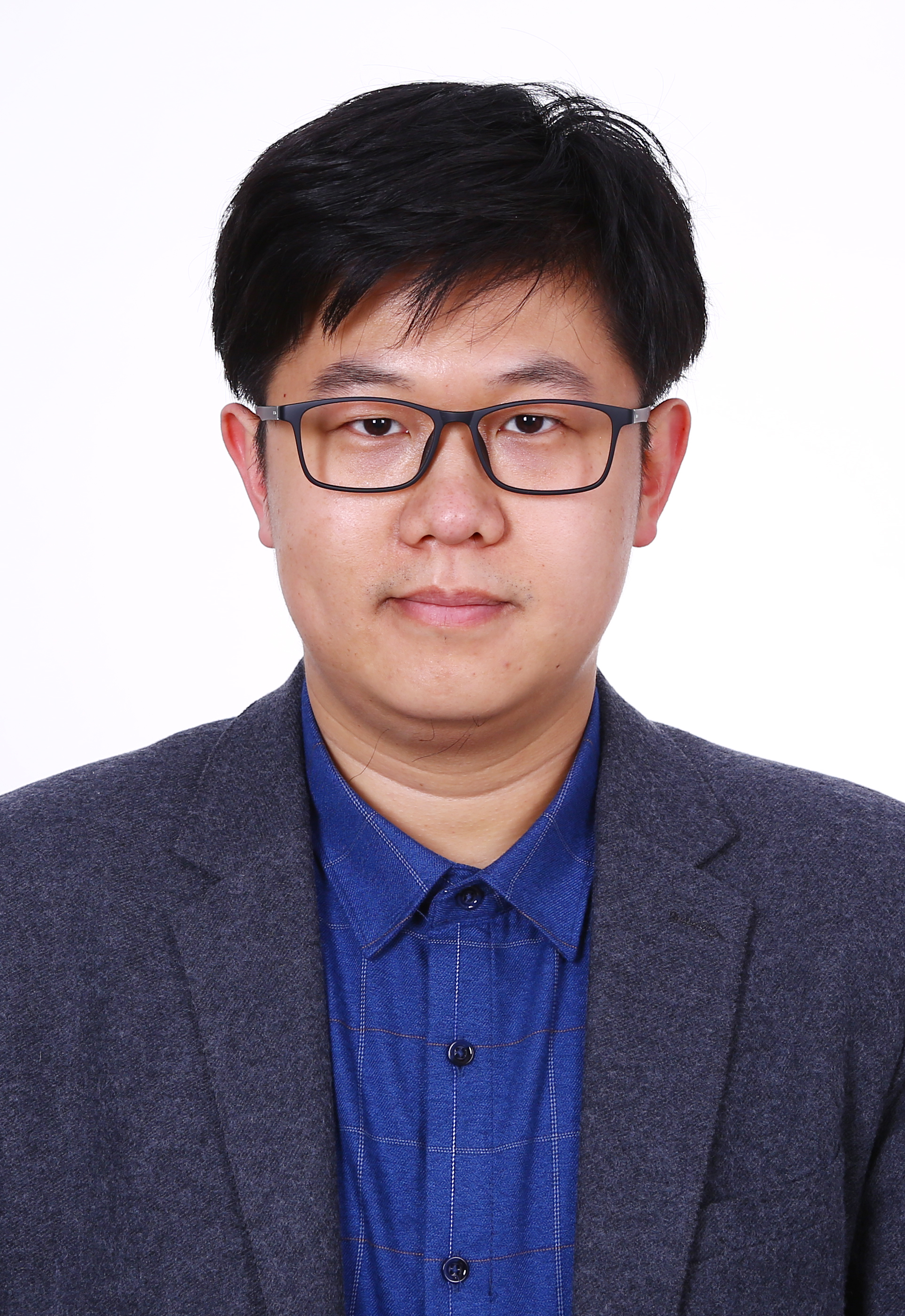}}]{Zhengxia Zou}
(Senior Member, IEEE) received his BS degree and his Ph.D. degree from Beihang University in 2013 and 2018. He is currently a Professor at the Department of Aerospace Intelligent Science and Technology, School of Astronautics, Beihang University. During 2018-2021, he was a postdoc research fellow at the University of Michigan, Ann Arbor. His research interests include computer vision and related problems in remote sensing. He has published over 30 peer-reviewed papers in top-tier journals and conferences, including Proceedings of the IEEE, Nature Communications, IEEE Transactions on Pattern Analysis and Machine Intelligence, IEEE Transactions on Geoscience and Remote Sensing, and IEEE / CVF Computer Vision and Pattern Recognition. Dr. Zou serves as the Associate Editor for IEEE Transactions on Image Processing. His personal website is \url{https://zhengxiazou.github.io/}.
\end{IEEEbiography}

\begin{IEEEbiography}[{\includegraphics[width=1in,height=1.25in,clip,keepaspectratio]{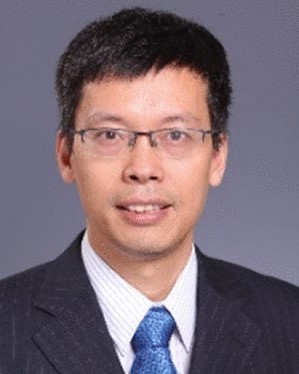}}]{Xiangyang Ji}(Member, IEEE) received the BE degree in materials science and the MS degree in computer science from the Harbin Institute of Technology, Harbin, China, in 1999 and 2001, respectively, and the PhD degree in computer science from the Institute of Computing Technology, Chinese Academy of Sciences, Beijing, China. He joined Tsinghua University, Beijing, in 2008, where he is currently a professor with the Department of Automation, School of Information Science and Technology. He has authored more than 200 refereed conference and journal papers. His current research interests include signal processing, computer vision, and computational photography.
\end{IEEEbiography}

\begin{IEEEbiography}[{\includegraphics[width=1in,height=1.25in,clip,keepaspectratio]{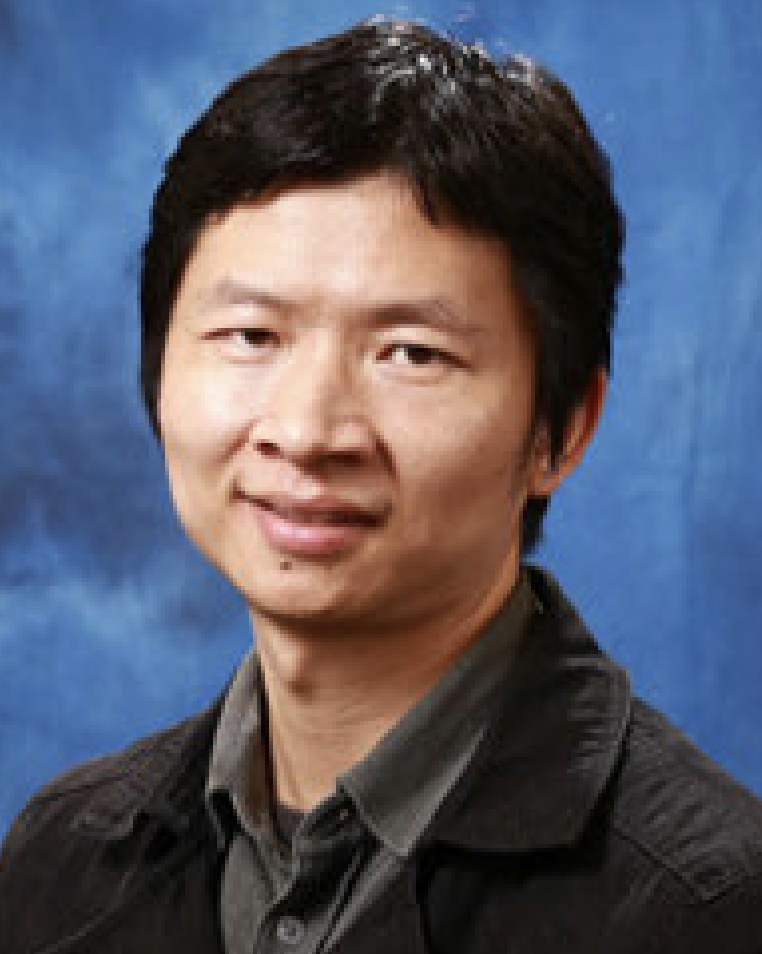}}]{Wanli Ouyang}
received the Ph.D. degree from the Department of Electronic Engineering, Chinese University of Hong Kong, Hong Kong, in 2010.

He was an Associate Professor with The University
of Sydney, Camperdown, NSW, Australia. He is now
a Professor with Shanghai Artificial Intelligence (AI) Laboratory, Shanghai, China. His research interests include pattern recognition, machine learning, and AI for Science.

Dr. Ouyang served as an Associate Editor for International Journal of Computer Vision (IJCV) and
Pattern Recognition (PR), the Senior Area Chair for Conference on Computer Vision and Pattern Recognition (CVPR), and the Guest Editor for IEEE
TRANSACTIONS ON PATTERN ANALYSIS AND MACHINE INTELLIGENCE (TPAMI).
\end{IEEEbiography}

\begin{IEEEbiography}
[{\includegraphics[width=1in,height=1.25in,clip,keepaspectratio]{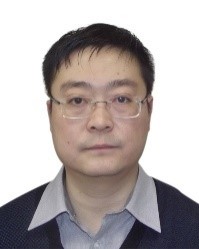}}]
{Zhenwei Shi}
(Senior Member, IEEE) is currently a Professor and Dean of the Department of Aerospace Intelligent Science and Technology, School of Astronautics, Beihang University. He has authored or co-authored over 200 scientific articles in refereed journals and proceedings, including the IEEE Transactions on Pattern Analysis and Machine Intelligence, the IEEE Transactions on Image Processing, the IEEE Transactions on Geoscience and Remote Sensing, the IEEE Conference on Computer Vision and Pattern Recognition (CVPR) and the IEEE International Conference on Computer Vision (ICCV). His current research interests include remote sensing image processing and analysis, computer vision, pattern recognition, and machine learning.

Prof. Shi serves as an Editor for IEEE Transactions on Geoscience and Remote Sensing, Pattern Recognition, ISPRS Journal of Photogrammetry and Remote Sensing, Infrared Physics and Technology, etc. His personal website is http://levir.buaa.edu.cn/.
\end{IEEEbiography}

\end{document}